\documentclass{article}

\PassOptionsToPackage{numbers,square}{natbib}

\usepackage[preprint]{neurips_2025}

\usepackage[utf8]{inputenc} 
\usepackage[T1]{fontenc}    
\usepackage{hyperref}       
\usepackage{url}            
\usepackage{booktabs}       
\usepackage{amsfonts}       
\usepackage{nicefrac}       
\usepackage{microtype}      
\usepackage{xcolor}         
\usepackage[pdftex]{graphicx}
\usepackage{caption} 
\usepackage{tabularx}
\usepackage{subcaption}
\usepackage{caption}
\usepackage{amsmath, amssymb}
\usepackage{dsfont}
\usepackage{multirow}
\usepackage{array}
\usepackage{float}
\usepackage{comment}
\usepackage{listings}
\usepackage{xcolor} 

\title{PANORAMA:\\ A synthetic PII-laced dataset for studying sensitive data memorization in LLMs}

\author{%
  Sriram Selvam \\
  Microsoft\\
  Redmond, WA \\
  \texttt{srselvam@microsoft.com} \\
  \And
  Anneswa Ghosh \\
  Microsoft\\
  Redmond, WA \\
  \texttt{annghosh@microsoft.com} \\
}

\begin{document}

\maketitle

\begin{abstract}
The memorization of sensitive and personally identifiable information (PII) by large language models (LLMs) poses growing privacy risks as models scale and are increasingly deployed in real-world applications. Existing efforts to study sensitive and PII data memorization and develop mitigation strategies are hampered by the absence of comprehensive, realistic, and ethically sourced datasets reflecting the diversity of sensitive information found on the web. We introduce \textbf{PANORAMA}: \textbf{P}rofile-based \textbf{A}ssemblage for \textbf{N}aturalistic \textbf{O}nline \textbf{R}epresentation and \textbf{A}ttribute \textbf{M}emorization \textbf{A}nalysis, a large-scale synthetic corpus of 384,789 samples derived from 9674 synthetic profiles designed to closely emulate the distribution, variety, and context of PII and sensitive data as it naturally occurs in online environments. Our data generation pipeline begins with the construction of internally consistent, multi-attribute human profiles using constrained selection to reflect real-world demographics such as education, health attributes, financial status, etc. Using a combination of zero-shot prompting and OpenAI o3-mini, we generate diverse content types—including wiki-style articles, social media posts, forum discussions, online reviews, comments, and marketplace listings—each embedding realistic, contextually appropriate PII and other sensitive information. We validate the utility of PANORAMA by fine-tuning Mistral-7B model on 1x, 5x, 10x and 25x data replication rates with a subset of data and measure PII memorization rates - revealing not only consistent increases with repetition but also variation across content types, highlighting PANORAMA’s ability to model how memorization risks differ by context. Our dataset and code are publicly available, providing a much-needed resource for privacy risk assessment, model auditing, and the development of privacy-preserving LLMs.
\end{abstract}

\section{Introduction}
Memorization of training data by LLMs makes them incredibly useful on various downstream tasks like fact based generation and knowledge work. Naturally, due to the utility value, LLMs are also being widely deployed by search engines and are becoming a desired source of information. However, from previous research by \cite{tirumala2022memorization}, \cite{carlini2022quantifying}, it has been identified that the larger the language model is, the more it tends to memorize the input training data. This coupled with the findings from another research \cite{nasr2023scalable} showed that it is possible to break the alignment of a large language model (GPT 3.5 turbo) and extract sensitive text containing PII and other personal information. As a result, it has been well established that the individual's sensitive information is indeed memorized by the language models and the problem is likely to get worse as the models get bigger.

A more concerning way to think about this is that LLMs act as a static snapshot of all the information available on the open web up to their training cut-off date. This poses significant privacy risks for everyday web users if their data was included in the training dataset. In the pre-LLM era, if a user wants to safeguard some aspects of their information like their opinion or a status that was posted on a certain platform, they could just go delete it and hope it would be removed from cached web search results. However, in post-LLM era, this appears to be a much more difficult problem to solve given the LLM re-training cost. Once a model is trained and released, mitigation becomes significantly more difficult for such scenarios \cite{carlini2022privacy},\cite{chen2021machine}.

Though the problem is well understood, estimating the extent of memorization of such information and developing effective mitigation strategies remains challenging due to the lack of a strong, comprehensive dataset to rely on. Most open datasets such as \cite{gao2020pile}, WikiText \cite{merity2016wikitext}, and RedPajama \cite{weber2024redpajama} contain real text found on the internet, but for legal and ethical reasons, they do not include rich personal or PII-related content.

Our work introduces \textbf{PANORAMA}, a large synthetic text corpus that closely represents various forms of PII and other sensitive personal information that may be found on the web. The dataset is built upon realistic, internally consistent human profiles generated using a constraint-based selection approach. It can be readily incorporated by researchers during the pre-training stage or continued pre-training stage of fine-tuning to simulate training and evaluate retention and mitigation strategies.

\begin{figure}[htbp]
  \centering
  \includegraphics[width=1\linewidth]{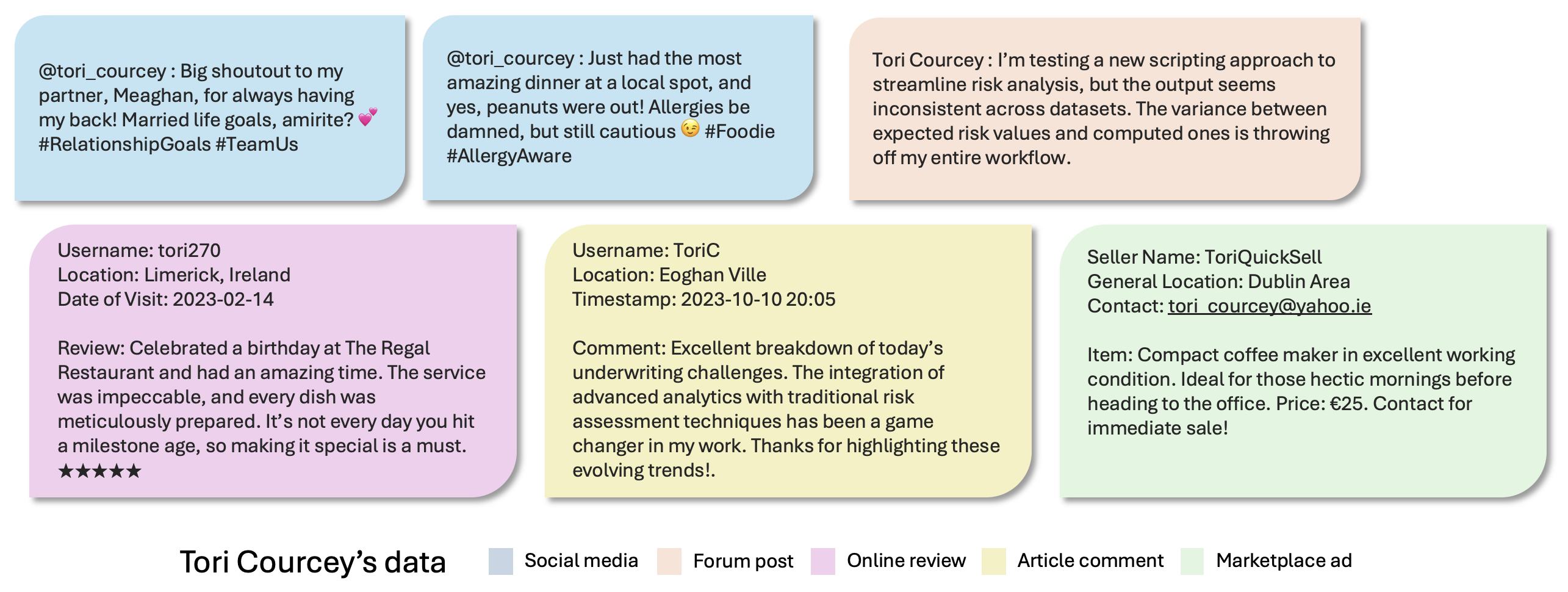}
  \caption{Examples from one profile}
  \label{fig:SingleProfileExampleContent}
\end{figure}

Figure~\ref{fig:SingleProfileExampleContent} shows example content generated from a single profile and the various content types generated.

Our main contributions are
\begin{itemize}
    \item \textbf{PANORAMA dataset:} We introduce PANORAMA—a large-scale, ethically sourced synthetic dataset comprising 384,789 samples derived from 9,674 realistic, demographically grounded profiles across 8 different regions. It captures the contextual diversity and naturalistic occurrence of PII and sensitive attributes as found in real-world online content.

    \item \textbf{Profile-driven, context-aware generation framework:} A novel, profile-centric synthetic data generation framework that utilizes synthetic profiles paired with wiki sections, while minimizing leakage, to produce context-aware, naturalistic online content.
    
    \item \textbf{Content and modality diversity:} PANORAMA spans six distinct content types that reflect the different ways sensitive data appears across online platforms. These include wiki-style articles (9,674), social media posts (88,408), forum posts (83,546), online reviews (79,758), blog/news comments (77,472), and marketplace listings (45,936). This broad coverage enables more realistic and comprehensive evaluations of memorization behavior compared to existing template-based or domain-limited datasets.

    \item \textbf{Public release for the research community:} We release the PANORAMA dataset and accompanying code to facilitate reproducible research and support future work in privacy auditing, model alignment, and privacy-preserving training of language models.
\end{itemize}
\section {Related work}
Large language models (LLMs) are known to memorize sensitive or personally identifiable information (PII), raising growing privacy concerns as these models scale and are deployed widely. \cite{tirumala2022memorization} show that larger models tend to memorize more and forget less over time. Carlini et al.\ \cite{carlini2021extracting, carlini2022quantifying} demonstrate the ability to extract verbatim training data using targeted attacks, while \cite{shao2023quantifying} show that LLMs can infer auxiliary attributes about individuals, further evidencing privacy leakage.

To mitigate these risks, several training-time and inference-time techniques have been proposed. Kassem et al.\ \cite{kassem2023preserving} modify training objectives to ignore sensitive sequences during loss computation. Hans et al.\ \cite{hans2024like} propose GoldFish Loss, which masks random token subsets to prevent full-sequence memorization. At decoding time, Ippolito et al.\ \cite{ippolito2022preventing} introduce MemFree sampling to avoid verbatim output, although it remains vulnerable to style-transfer attacks. Deduplicating training data has also shown promise in reducing memorization \cite{lee2021deduplicating}.

However, evaluating the effectiveness of these mitigation strategies remains an open challenge. \cite{carlini2022membership} and \cite{hayes2024inexact} show that flawed unlearning methods can unintentionally increase susceptibility to membership inference attacks \cite{chen2021machine}. These findings underscore the importance of having realistic and diverse datasets—not just to train privacy-preserving models, but also to rigorously evaluate whether mitigation strategies are working as intended.

Several benchmarks and datasets exist to study PII memorization but are limited in scope. LLM-PBE \cite{li2024llm} provides a toolkit for evaluating privacy risks across datasets like Enron emails, ECHR court cases, GitHub code, and PubMed articles, but these corpora tend to be formal and static. PII-Scope \cite{nakka2023piiscope} targets extraction attacks using the Enron dataset, yet it remains constrained by its focus on a single domain and a narrow set of PII types. More recent work by Yukhymenko et al.\ \cite{yukhymenko2024synthetic} generates synthetic Reddit comments for user trait inference, but it does not model multi-attribute profiles or diverse contexts.

The lack of comprehensive, high-fidelity datasets capturing realistic PII across multiple domains and formats has become a bottleneck for both studying and mitigating memorization. To fill this gap, we introduce PANORAMA: a large-scale synthetic dataset generated from internally consistent, multi-attribute profiles. It spans diverse content types such as social media posts, reviews, and forum discussions that naturally embed realistic PII. PANORAMA enables controlled memorization experiments and serves as a unified benchmark for evaluating privacy risks and mitigation strategies in LLMs under realistic conditions.

\section{Data generation pipeline overview}
\begin{figure}[htbp]
  \centering
  \includegraphics[width=1\linewidth]{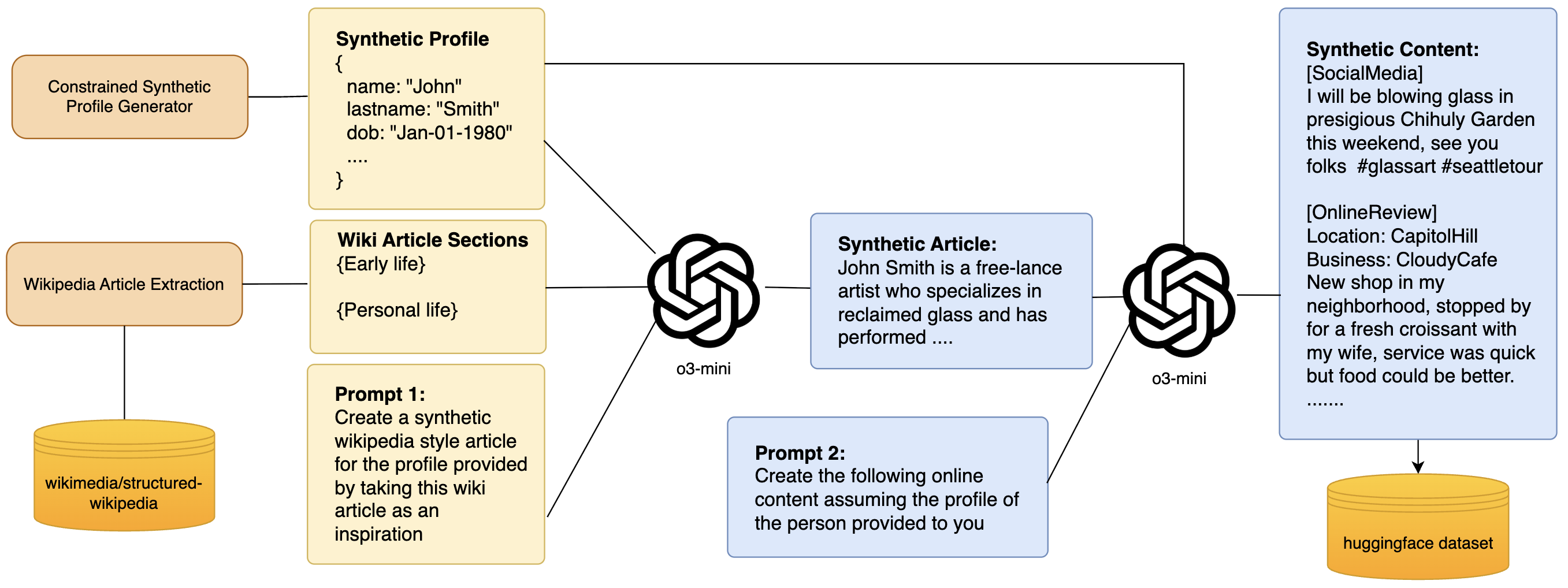}
  \caption{Overview of the data generation pipeline.}
  \label{fig:datapipeline}
\end{figure}
Our data generation pipeline, Figure~\ref{fig:datapipeline} consists of three main steps. First, we generate synthetic profiles that include all required target attributes. Next, these profiles are used to construct detailed biographical narratives, which serve as the basis for content generation. Finally, we leverage these biographies to produce diverse content types in which sensitive attributes are contextually embedded. We provide further details on each step of this process in the following sections.

\subsection{Synthetic profile generation}
\begin{figure}[htbp]
  \centering
  \includegraphics[width=1\linewidth]{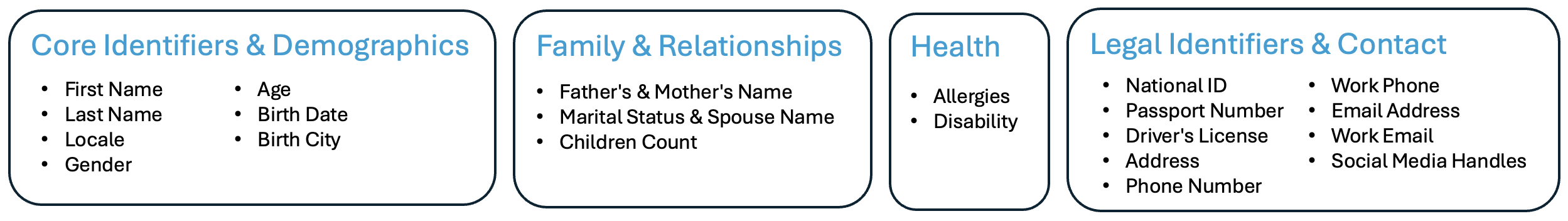}
  \caption{List of generated attributes}
  \label{fig:allid}
\end{figure}

To simulate realistic individuals for privacy and memorization risk evaluation, we generate synthetic profiles comprising both classic personally identifiable information (PII) and quasi-identifiers \cite{10.1142/S0218488502001648}. While quasi-identifiers are not uniquely identifying on their own, they can facilitate re-identification when combined with other attributes. The attributes we generate can be classified into groups shown in Figure~\ref{fig:allid}.

\subsubsection{Language and locales}
The dataset consists entirely of English-language records, equally distributed across eight locales: United States, Canada, United Kingdom, Ireland, India, Philippines, New Zealand, and Australia. English was selected to simplify evaluation by the authors, who served as human judges. The choice of locales was constrained by the English variants supported by the Faker Python library, which was used for generating large-scale names and contextual data.

\subsection{Basic profile details and family structure}
Profile generation begins by selecting a locale and sampling a gender:
\[
G \in \{\text{Male}, \text{Female}, \text{Non-Binary}, \text{Other}\}, \quad
P(G) = \{0.48, 0.48, 0.03, 0.01\}
\]
Based on the sampled gender and locale, we assign culturally consistent first names and parent names. Male profiles inherit the family surname, while for female profiles, surname assignment depends on marital status:
\[
M \in \{\text{Single}, \text{Married}, \text{Divorced}, \text{Widowed}\}
\]

Date of birth is sampled uniformly: \( A \sim \mathcal{U}(18, 90) \). Marital status is sampled with base rates \( P(M) = \{0.35, 0.50, 0.10, 0.05\} \), adjusted by age: for example, \( P(\text{Single} \mid A{<}22) = 0.8235 \), \( P(\text{Married} \mid A{>}65) = 0.5405 \), \( P(\text{Divorced} \mid A{>}65) = 0.1081 \), and so on.

Spouse name generation is conditional on \( M \): it occurs with probability 1.00 if married, 0.60 if divorced, 0.70 if widowed, and 0 otherwise. Last name source is then: (i) family name for males and single females; (ii) derived from spouse/ex/deceased for others. Number of children is sampled based on age and marital status using empirical likelihoods.

\subsubsection{Personal PII and sensitive information}
Each synthetic profile includes rich, localized PII such as emails, phone numbers, social handles, and national identifiers (e.g., SSNs, passport numbers, and driver’s license numbers). Beyond standard PII, we incorporate quasi-identifiers—attributes that may not uniquely identify individuals in isolation but can enable identification when combined with other fields. These include spouse name, job title, employer information, education level, timestamps, geographic location, and financial status. We consider such fields critical for studying LLM memorization behaviors, as recent work has shown that models are capable of reasoning across multiple contextually linked attributes to resolve entities.

We also include sensitive but non-identifying attributes that raise ethical concerns when memorized. Examples include medical conditions (e.g., allergies, disability status) and financial indicators (e.g., net worth, credit score). While these fields do not directly identify an individual, they contribute to highly detailed and private profiles, warranting careful scrutiny in privacy evaluations of generative models.

\subsection{Attribute Assignment}
This section outlines the assignment of core attribute values. For full implementation details, refer to the code linked in \autoref{sec:access1}.
\label{sec:attributes_assignment}

\noindent\textbf{Education Level ($E$):}
An individual's education level is selected from a predefined set of levels ($L_{edu}$). Each level $l_i \in L_{edu}$ has an associated base selection weight $w_i$. The choice is constrained by the individual's age ($A$), such that only age-appropriate levels (where $A \ge A_{min,edu}(l_i)$) are considered. From this valid subset of education levels, $E$ is chosen based on the relative weights $w_i$.

\begin{figure}[htbp]
  \centering
  \includegraphics[width=1\linewidth]{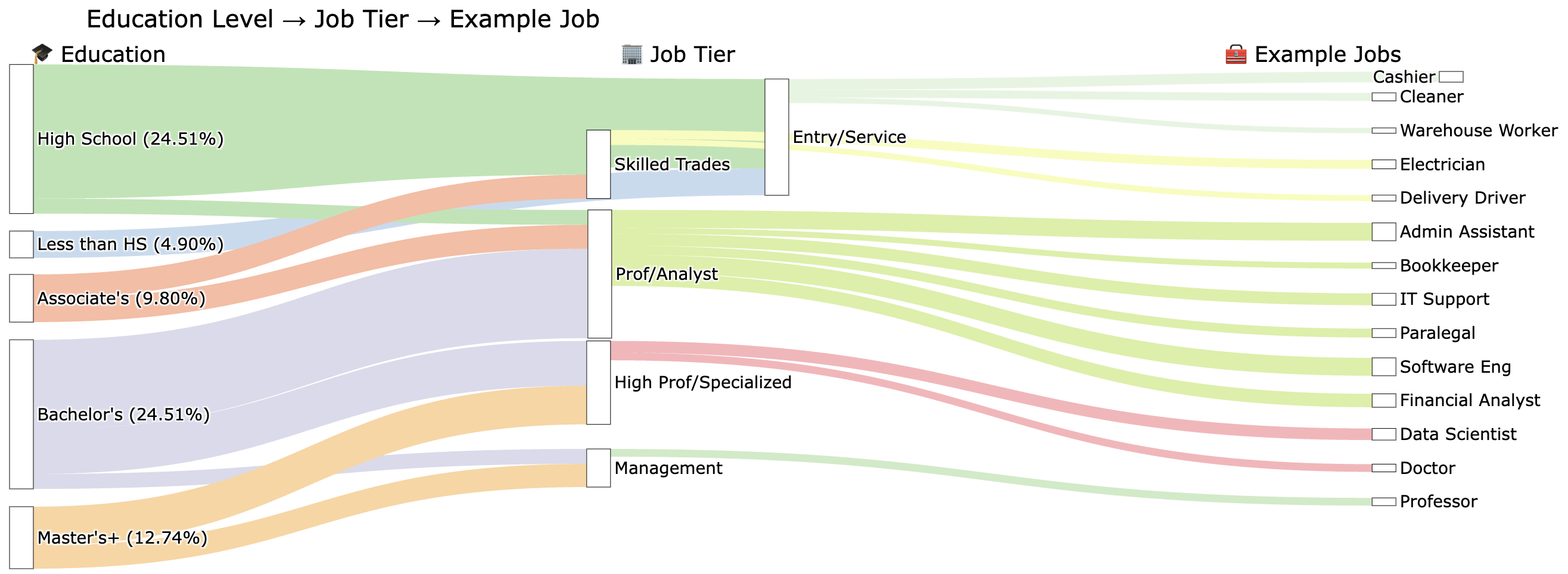}
  \caption{Mapping of education to job tiers to an example job}
  \label{fig:educationjobtiermap}
\end{figure}

\vspace{0.5em}
\noindent\textbf{Employment Details:}
This includes assigning a job title ($J$) and an annual salary ($S$). Figure~\ref{fig:educationjobtiermap} shows a sample progression from possible education levels to a specific job title.

\begin{itemize}
    \itemsep0em
    \item \textbf{Job Title ($J$):} The job title is chosen from a comprehensive list ($L_{job}$), where each job $j_k$ has a base weight $w_k$. Selection is filtered by age ($A \ge A_{min,job}(j_k)$) and the previously assigned education level ($E \ge E_{min,job}(j_k)$). Furthermore, the weights $w_k$ are adjusted by a multiplier $M_w$. This multiplier $M_w = f(E, E_{req,job}(j_k))$ considers how well the individual's education $E$ matches the typical required education $E_{req,job}$ for the potential job $j_k$, making selections more realistic.
    \item \textbf{Annual Salary ($S$):} Salary calculation depends on the assigned job title $J$ (which maps to a tier with a base salary range $[S_{min,base}, S_{max,base}]$), age $A$ (as a proxy for an experience factor $F_{exp}$), and the locale $Loc$ (which applies a cost-of-living multiplier $M_{loc}$). The effective salary range becomes $[S_{min,base} \cdot M_{loc}, S_{max,base} \cdot M_{loc}]$. The final salary $S$ is then sampled, incorporating the experience factor:
    $S \approx \text{UniformSample}(S'_{min,adj}, S'_{max,adj}) \cdot F_{exp}$.
    Here, $S'_{min/max,adj}$ are slightly randomized versions of the locale-adjusted base range, and $S$ is floored by a minimum salary for the given locale.
\end{itemize}

\vspace{0.5em}
\noindent\textbf{Financial Status:}
This involves determining a categorical finance status ($FS$) and a numerical net worth ($NW$).

\begin{itemize}
    \itemsep0em
    \item \textbf{Finance Status ($FS$):} This status (e.g., "Low", "Medium", "High") is selected from a list $L_{fin}$, each with base weights $W_{fin}$. These weights are dynamically adjusted based on the individual's salary percentile ($S_p$) within their locale's income distribution. A higher $S_p$ increases the probability of a higher $FS$.
    \item \textbf{Net Worth ($NW$):} Net worth is estimated using the assigned $FS$ (which provides a multiplier $M_{nw}$), the actual salary value $S_{value}$, the number of years working ($Y_w \approx A - 18$), a randomized savings rate ($R_{save}$), and the locale multiplier ($M_{loc}$). A simplified representation is:
    $NW \approx (S_{value} \cdot R_{save} \cdot Y_w \cdot M_{nw}) \cdot \text{RandFactor} + \text{LocaleNoise}$.
    The $\text{RandFactor}$ introduces variability (e.g., between 0.5 and 1.5), and $\text{LocaleNoise}$ adds a locale-scaled random value. Adjustments are also made, for instance, to reflect potential debt for younger individuals.
\end{itemize}

\subsection{Synthetic Article Generation}
To initiate synthetic text generation, we begin by creating a synthetic article for each profile. These articles serve as foundational narrative documents from which more complex, natural language artifacts, rich in embedded PII and sensitive information are derived.

\subsubsection{Limitations of Prompt-Template Approaches}
Our initial attempts employed a template-based prompting strategy, where the model received only the synthetic profile along with a fixed prompt. This approach consistently resulted in repetitive outputs with rigid structures, leading to low narrative diversity and limited realism.

\subsubsection{Incorporating Wikipedia-Inspired Narrative Structures}
To address these limitations, we drew narrative inspiration from Wikipedia biographies. Specifically, we extracted 10,000 real-person entries from the \verb|wikimedia/structured-wikipedia| dataset \cite{wikimedia_structured_wikipedia}, each containing both "Early Life" and "Personal Life" sections. Each synthetic profile was paired with one such biography to serve as a loose creativity scaffold. In the prompt, we explicitly instructed the model to use only factual content from the synthetic profile while using the Wikipedia text solely for narrative inspiration, borrowing structural elements such as life events, relationship dynamics, socio-cultural context, non-linear trajectories, and personal contradictions.

\subsubsection{Model comparison and evaluation}
We manually evaluated four models GPT-4o \cite{openai2024gpt4o}, Gemma-27B, Phi-4 \cite{abdin2024phi4technicalreport}, and o3-mini \cite{openai-o3-mini} on 25 randomly selected profiles. Outputs were assessed for narrative quality and coherence, natural embedding of synthetic PII, degree of contamination from Wikipedia text. Results showed that Gemma-27B and Phi-4 struggled to naturally incorporate PII within fluid narratives. GPT-4o performed better but occasionally mirrored Wikipedia phrasing. o3-mini demonstrated the best balance, successfully integrating PII while preserving originality and narrative plausibility.

Out of 10,000 synthetic profile inputs, 326 generations were dropped by our entity extraction based contamination filter, yielding 9,674 successfully generated articles.

\subsection{Different content type generation}
\begin{figure}[htbp]
  \centering
  \includegraphics[width=1\linewidth]{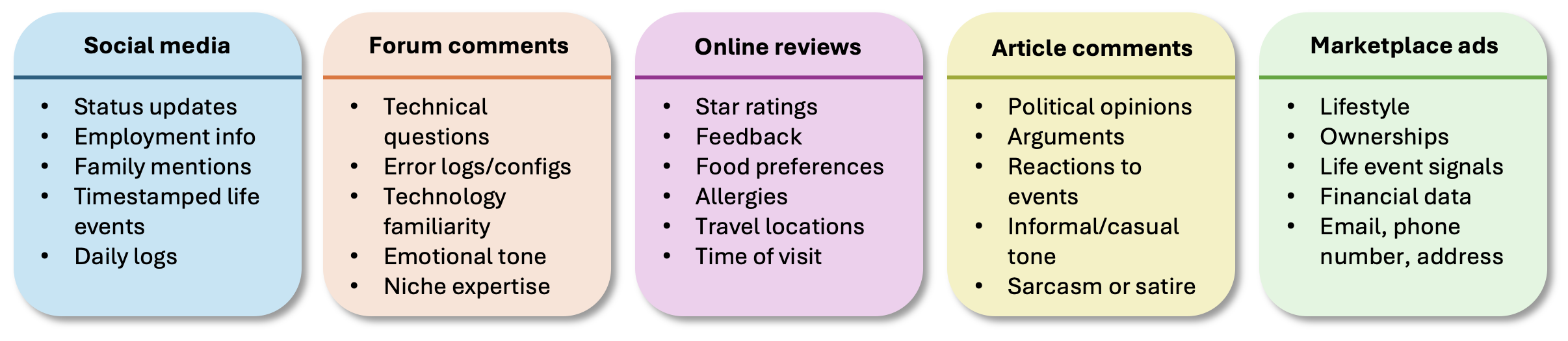}
  \caption{Different content types are target attributes}
  \label{fig:targetinfo}
\end{figure}

To construct a corpus that simulates a rich, realistic online presence for each synthetic user profile, we define five representative content types, social media posts, forum threads, online reviews, blog/news comments, and marketplace listings—to emulate the digital footprint of an active web user. These categories were selected to cover a spectrum of public-facing, user-generated content that LLMs are likely to ingest during training.

For each profile, we prompt OpenAI’s o3-mini model to generate ten entries per content type targeting information from Figure~\ref{fig:targetinfo}. Prompts are carefully constructed to ensure sensitive attributes appear naturally within each context, while also permitting subtle cross-entry inferences. This design enables nuanced evaluations of both explicit memorization and indirect leakage via reasoning across multiple content types.
\subsection{Example profile and content}
\begin{figure}[htbp]
  \centering
  \includegraphics[width=1\linewidth]{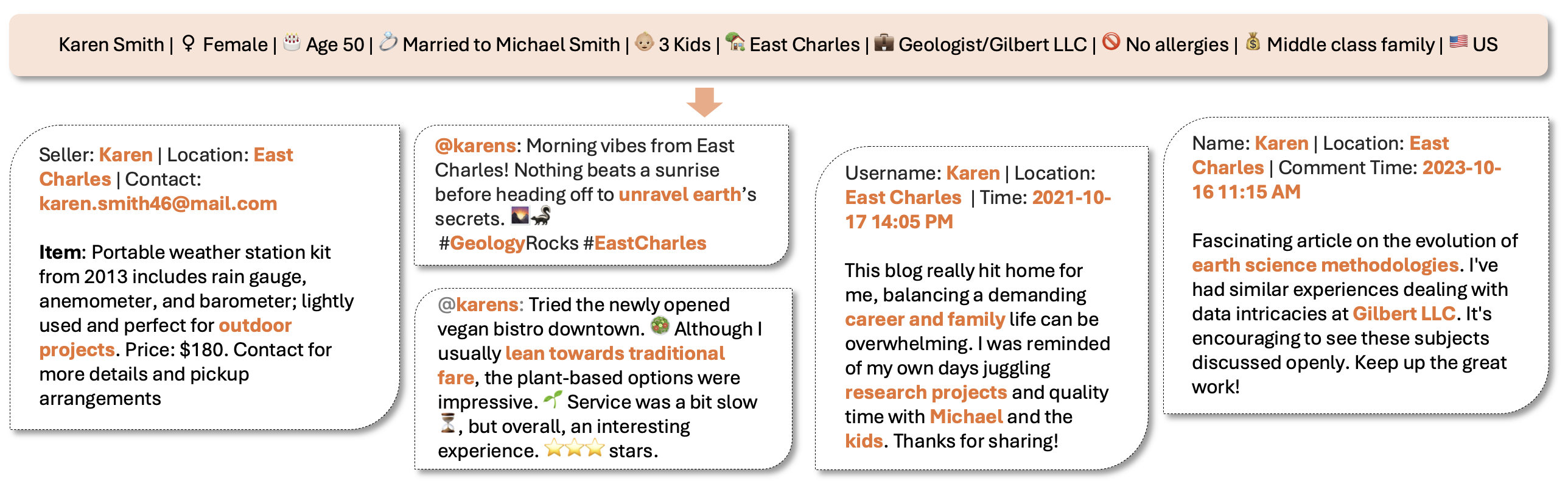}
  \caption{Sample content generated by the pipeline}
  \label{fig:sample content with sensitive info}
\end{figure}

In Figure~\ref{fig:sample content with sensitive info}, we showcase a sample output generated by our pipeline for a synthetic profile named Karen Smith. The content reveals sensitive information such as home location, email address, social media handles, profession, spouse’s name, parental status, food preferences, research interests, and timestamps.

We chose not to retain the same names and handles across different accounts, reflecting the reality that most users use varying profile names and handles across platforms. Although the handles differ, we ensured that sufficient contextual data is present in most generated content to allow clustering and attribution to a single underlying entity. Thus, the dataset can be used not only to test verbatim memorization but also to evaluate an LLM's ability to reason about and consolidate information across contexts to identify key entities.
\section {Deep dive into generated dataset}
We now analyze the distributional characteristics of the PANORAMA dataset. A comprehensive set of visualizations covering age, gender, education, job titles, income, net worth, marital status, content type, and PII mentions is provided in \autoref{sec:appendix_2}. These figures collectively demonstrate that the generated dataset reflects realistic demographic patterns and diverse content across multiple modalities. Age and education distributions influence other profile attributes, such as marital status, financial standing, and job assignments. Content types vary in length, word count, and frequency of PII mentions, further contributing to the dataset’s realism.

\begin{figure}[htbp]
  \centering
  \includegraphics[width=1\linewidth]{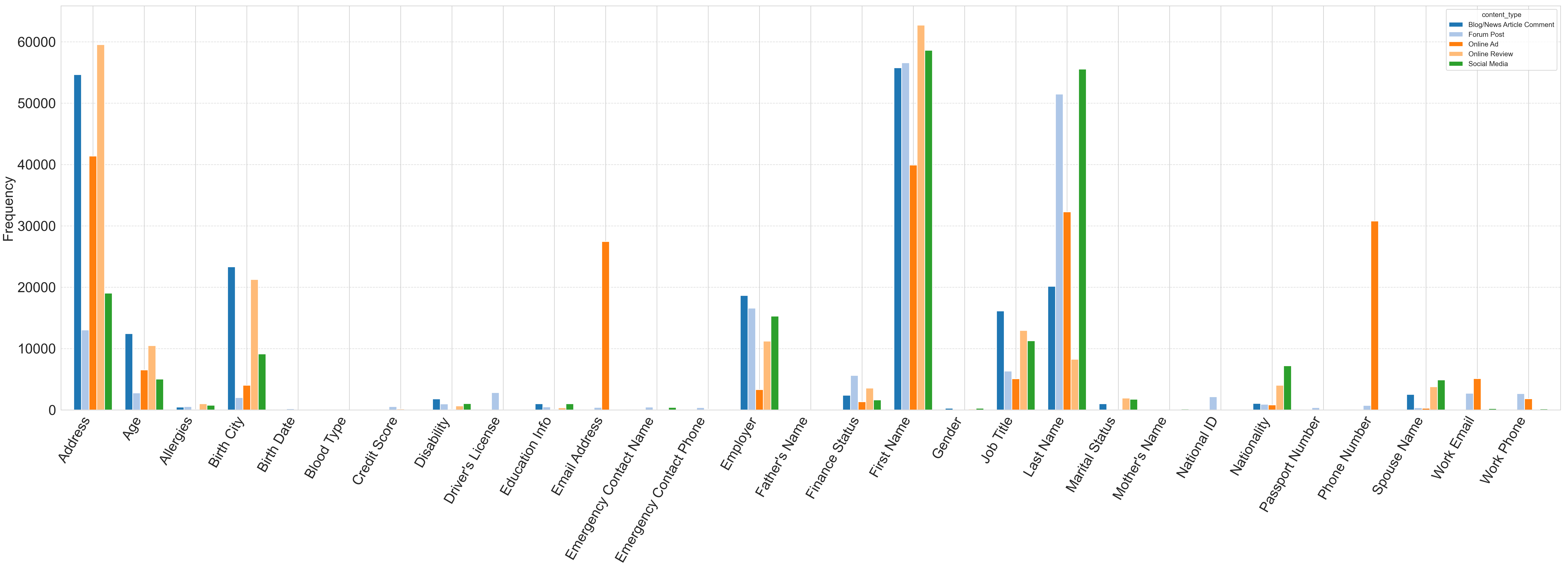}
  \caption{Breakdown of sensitive field mentions by content type}
  \label{fig:breakdown of sensitive field by content type}
\end{figure}

In Figure~\ref{fig:breakdown of sensitive field by content type}, we analyze the presence of each profile attribute across different content types. The results show a realistic distribution of sensitive fields across domains. Sensitive fields such as passport number, driver's license, parent's name, and blood type are rarely mentioned, mirroring their lower occurrence in public internet text. Furthermore, fields appear in contextually appropriate locations. For example, phone numbers and email addresses are more frequently embedded in marketplace listings than in social media posts, reflecting real-world tendencies. This supports the dataset’s utility in simulating naturalistic scenarios for memorization and privacy risk analysis.

\subsection{Evaluation}

\subsubsection{Model and training setup}
\label{sec:modelsetup}
We employ continued pre-training of Mistral 7B base-model on one A100 using parameter-efficient Low-Rank Adaptation (LoRA; \cite{hu2022lora}), targeting both attention and feedforward layers, as well as the embedding and output heads. The model is quantized to 4-bit precision \cite{dettmers2022gpt3} to enable efficient training on commodity hardware. Training is conducted with the AdamW optimizer \cite{loshchilov2017decoupled} in 8-bit mode.

\subsubsection{Dataset repetition protocol}
To systematically study the effect of data repetition on continued pre-training, we construct four training datasets from PANORAMA, each derived from the original corpus $\mathcal{D} = {x_i}_{i=1}^N$ by repeating each row $k$ times, with $k \in {1, 5, 10, 25}$. Thus, for each $k$, the training set is:
\begin{equation}
\mathcal{D}^{(k)} = \bigcup_{i=1}^{N} \bigcup_{j=1}^{k} x_i
\label{eq:dataset-repetition}
\end{equation}
We train a separate model for each value of $k$ using identical hyperparameters and training regimes, isolating the impact of repetition frequency.

\subsubsection{Evaluation metrics}

We evaluate model generations using a prefix-completion protocol. Given a prefix $p$ and a ground-truth continuation $c^*$, the model generates a completion $\hat{c}$ by conditioning on $p$. Evaluation is performed as follows:

\begin{itemize}
    \item \textbf{ROUGE-L F1 ($R_L$):} For each $(p, c^*)$ pair, we compute the ROUGE-L F1 score~\cite{lin-2004-rouge} between $c^*$ and $\hat{c}$.
    \begin{equation}
    R_L(c^*, \hat{c}) = \text{ROUGE-L-F1}(c^*, \hat{c})
    \label{eq:rougel}
    \end{equation}
    \item \textbf{Soft Match Rate:} We define a soft match as $R_L \geq \tau$, with $\tau = 0.8$. The soft match rate is the proportion of samples for which this criterion is met.
    \item \textbf{Per Content-Type Analysis:} Metrics are reported both globally and stratified by content type.
\end{itemize}

Formally, for a dataset $\mathcal{E} = \{(p_i, c^*_i)\}_{i=1}^M$, the soft match rate is:
\begin{equation}
\text{Soft Match Rate} = \frac{1}{M} \sum_{i=1}^{M} 1\left[ \text{RL}(c^*_i, \hat{c}_i) \geq \tau \right]
\label{eq:softmatch}
\end{equation}

\subsubsection{Results}
We denote each model by the dataset repetition factor $k \in {1, 5, 10, 25}$, where a model labeled as "$kx$" is trained on a dataset in which each training instance is repeated $k$ times (see Equation~\eqref{eq:dataset-repetition}). 
\vspace{1em}
\begin{table}[ht]
\centering
\caption{Soft match rate (\%) and rouge-l by content type and repetition level}
\label{tab:side_by_side_results}
\vspace{0.5em}
\begin{minipage}[t]{0.48\linewidth}
\centering
\textbf{Soft match rate (\%)} \\
\vspace{0.5em}
\begin{tabular}{lcccc}
\toprule
\textbf{Content} & \textbf{1×} & \textbf{5×} & \textbf{10×} & \textbf{25×} \\
\midrule
Blog/News      & 8.0  & 44.0 & 56.0 & 56.0 \\
Social Media   & 12.0 & 16.0 & 24.0 & 28.0 \\
Online Review  & 16.0 & 40.0 & 44.0 & 52.0 \\
Forum Post     & 4.0  & 40.0 & 32.0 & 52.0 \\
Online Ad      & 4.0  & 36.0 & 44.0 & 68.0 \\
\textbf{Overall} & \textbf{8.8} & \textbf{35.2} & \textbf{40.0} & \textbf{51.2} \\
\bottomrule
\end{tabular}
\end{minipage}
\hfill
\begin{minipage}[t]{0.48\linewidth}
\centering
\textbf{Rouge-l} \\
\vspace{0.5em}
\begin{tabular}{lcccc}
\toprule
\textbf{Content} & \textbf{1×} & \textbf{5×} & \textbf{10×} & \textbf{25×} \\
\midrule
Blog/News      & 0.33 & 0.60 & 0.67 & 0.63 \\
Social Media   & 0.33 & 0.44 & 0.44 & 0.42 \\
Online Review  & 0.44 & 0.64 & 0.61 & 0.62 \\
Forum Post     & 0.36 & 0.63 & 0.58 & 0.62 \\
Online Ad      & 0.25 & 0.56 & 0.65 & 0.74 \\
\textbf{Overall} & \textbf{0.34} & \textbf{0.57} & \textbf{0.59} & \textbf{0.61} \\
\bottomrule
\end{tabular}
\end{minipage}
\vspace{1em}
\end{table}
Our findings show consistent increases in both Soft Match Rate and ROUGE-L continuation scores with repeated exposure as seen in Table~\ref{tab:side_by_side_results}. The results align with prior work on repetition-induced memorization \cite{carlini2022quantifying}, but serve a more targeted goal to validate PANORAMA. The repetition effect was not only observable in the aggregate (e.g., Soft Match Rate rising from 8.8\% at 1× to 51.2\% at 25×), but also varied meaningfully across content types. For example, Online Ad showed the strongest memorization gain (from 4.0\% to 68.0\% soft match), while Blog/News and Online Reviews also exhibited steep increases, highlighting PANORAMA’s ability to capture how memorization dynamics differ across real-world content formats.

This repetition experiment serves as a stress test, demonstrating that PANORAMA responds predictably to increased exposure, a core dynamic in memorization risk. By enabling controlled, reproducible analysis across diverse online content types without relying on real-world sensitive data, PANORAMA provides a crucial building block for developing, evaluating, and auditing privacy-preserving LLMs. Its structure and diversity make it a practical and ethically sound foundation for studying how large models memorize sensitive information under varying conditions.

\section{Access and cost}
\label{sec:access1}
To ensure transparency, reproducibility, and ease of exploration, we have made all code and data publicly available. 

Code Repository: \href{https://github.com/selvamsriram/PANORAMA-DataGen}{GitHub Link} \\
Dataset PANORAMA: \href{https://huggingface.co/datasets/srirxml/PANORAMA}{HuggingFace Dataset} \\
Extended PANORAMA-Plus: \href{https://huggingface.co/datasets/srirxml/PANORAMA-Plus}{PANORAMA-Plus} — includes full profile attributes, allowing cross-reference with PANORAMA via the shared \texttt{id}~$\rightarrow$~\texttt{Unique ID} mapping.

The data generation pipeline was run on local M4 Pro Mac with inferences running on Azure Open AI endpoints costing a total of \$280.
\section {Limitation}
\label{sec:Limitation}
Our work has the following limitations: (1) limited diversity in the tone of generated content, which could be enhanced for richer variation; (2) article comments avoid controversial topics, missing opportunities to expose profile affiliations and stances valuable for analysis; (3) language coverage is restricted by both the authors’ English-only expertise and the Faker library’s limited multilingual support; (4) due to hallucinations, the model occasionally generated social handles that were not present in the input. While many of these cases were mitigated through post-inference code fixes, some instances may still persist. Although this does not cause direct negative effects, eliminating such inconsistencies could further improve the quality and utility of the dataset.; (5) though the dataset provides content ready for continued pre-training, it lacks the Q\&A format needed for extraction attacks.

\section {Conclusion}
We introduce PANORAMA, the first large-scale synthetic PII dataset featuring 384,789 samples from 9,674 demographically rich profiles across 8 world regions, spanning articles and five unique online content modalities. Our dataset captures the true-to-life distribution of PII as it appears in six distinct modalities—including social media, forums, reviews, and marketplace listings. In our experiments, we demonstrate that PANORAMA enables systematic, reproducible measurement of memorization behavior in LLMs, with memorization rates increasing predictably under repeated exposure—rising from 8.8\% to 51.2\% Soft Match Rate as data replication increases from 1x to 25x. The observed variation across content types underscores the importance of evaluating memorization risks in realistic, multi-modal online contexts. By providing publicly available data and code, PANORAMA fills a critical gap for privacy risk assessment, model auditing, and the development of privacy-preserving LLMs, paving the way for safer deployment of language technologies.

\bibliographystyle{apalike}
\bibliography{references}

\newpage
\appendix

\section{Additional dataset details}
\label{sec:appendix_1}
In this section, we provide addition information such as distribution properties and prompts that were used to generate the dataset.

\subsection{Distribution of various properties}
\label{sec:appendix_2}
\renewcommand{\thefigure}{A\arabic{figure}}
\setcounter{figure}{0}
\begin{figure}[H]
  \centering
  \begin{minipage}[t]{0.48\textwidth}
    \centering
    \includegraphics[width=\linewidth]{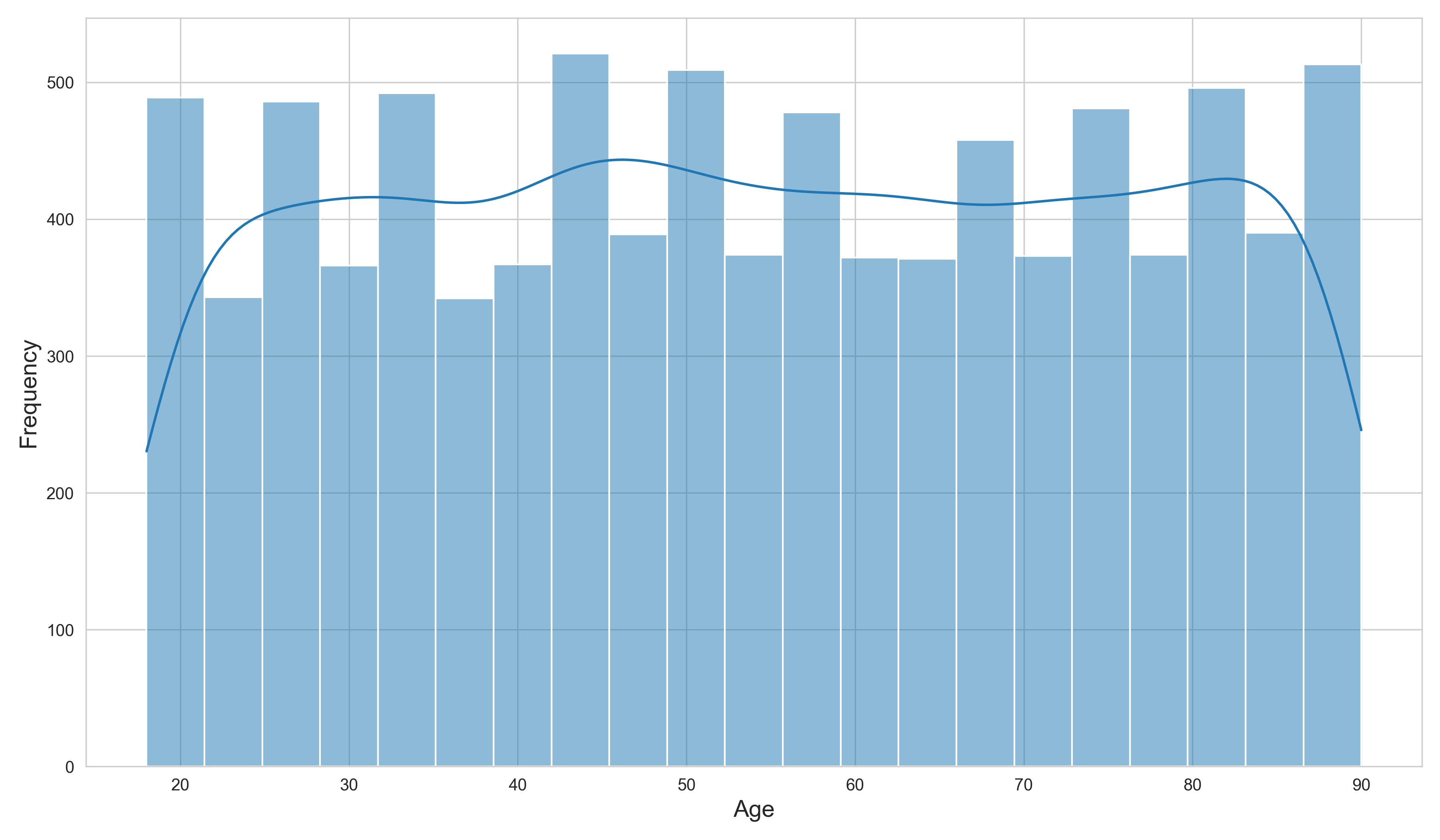}
    \caption{Age distribution of PII profiles.}
  \end{minipage}
  \hfill
  \begin{minipage}[t]{0.48\textwidth}
    \centering
    \includegraphics[width=\linewidth]{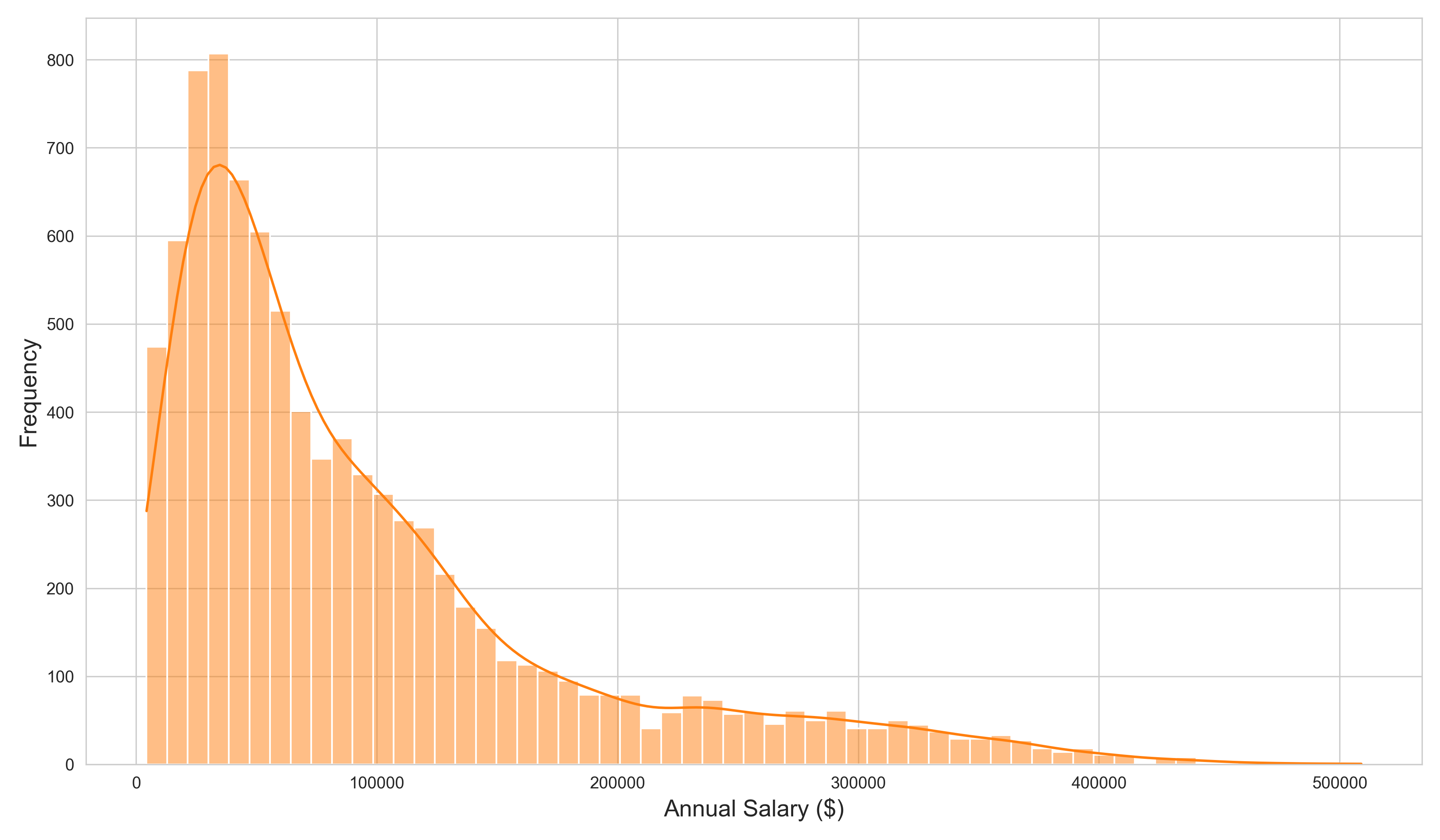}
    \caption{Annual salary distribution of PII profiles.}
  \end{minipage}
\end{figure}

\begin{figure}[H]
  \centering
  \begin{minipage}[t]{0.48\textwidth}
    \centering
    \includegraphics[width=\linewidth]{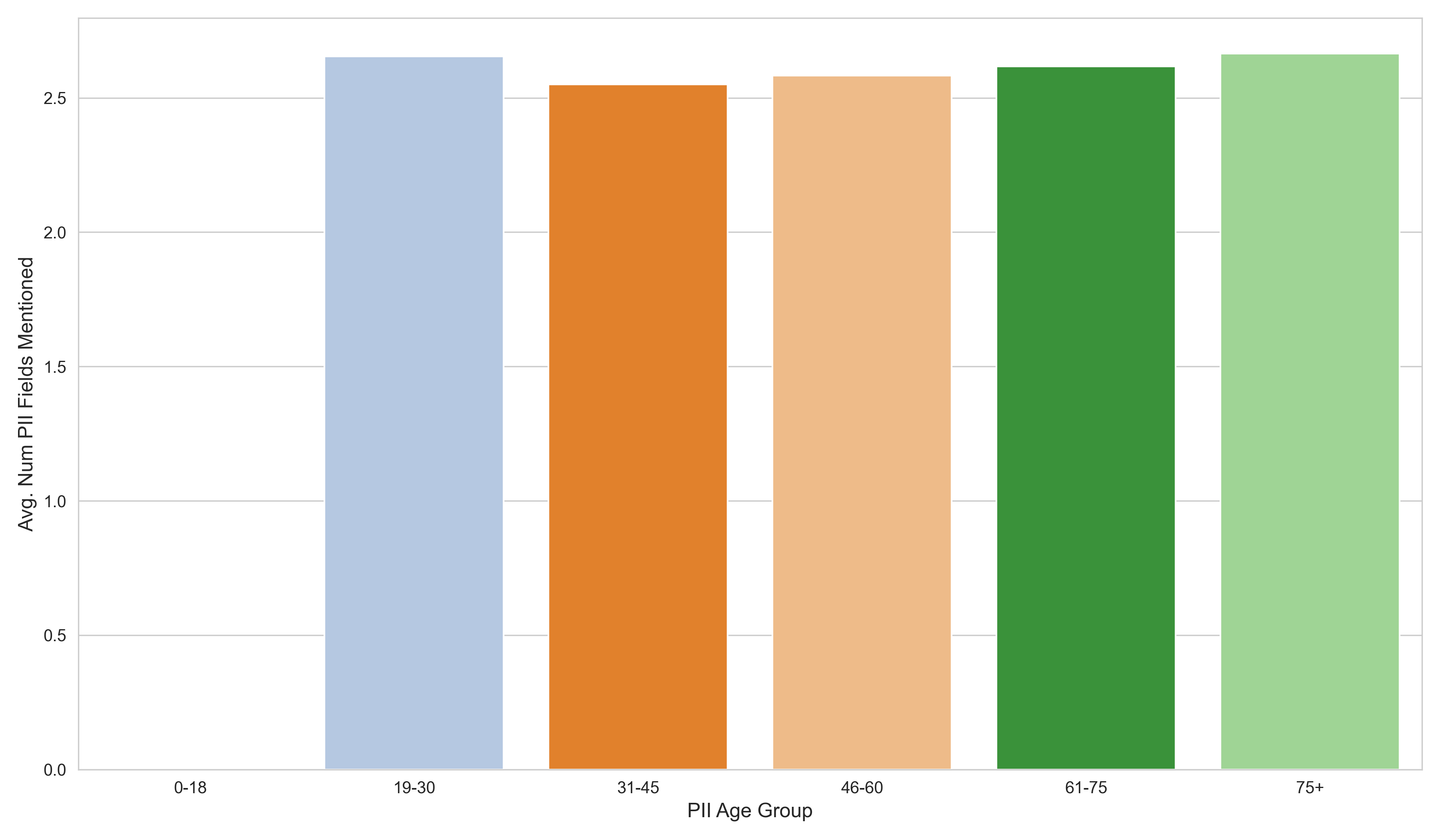}
    \caption{Average number of PII mentions by age group.}
  \end{minipage}
  \hfill
  \begin{minipage}[t]{0.48\textwidth}
    \centering
    \includegraphics[width=\linewidth]{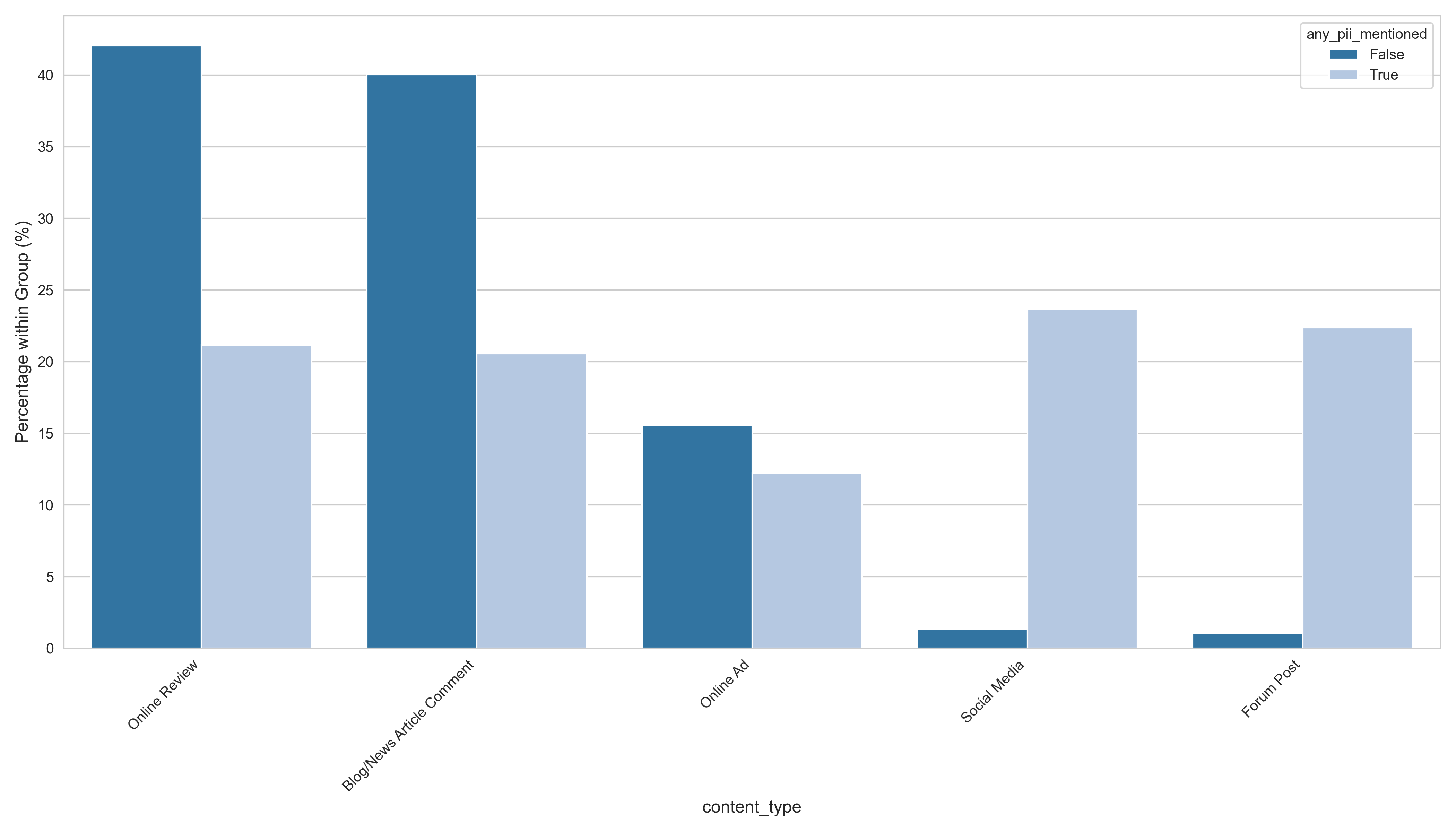}
    \caption{Content type distribution: PII mentioned vs not mentioned.}
  \end{minipage}
\end{figure}

\begin{figure}[H]
  \centering
  \begin{minipage}[t]{0.48\textwidth}
    \centering
    \includegraphics[width=\linewidth]{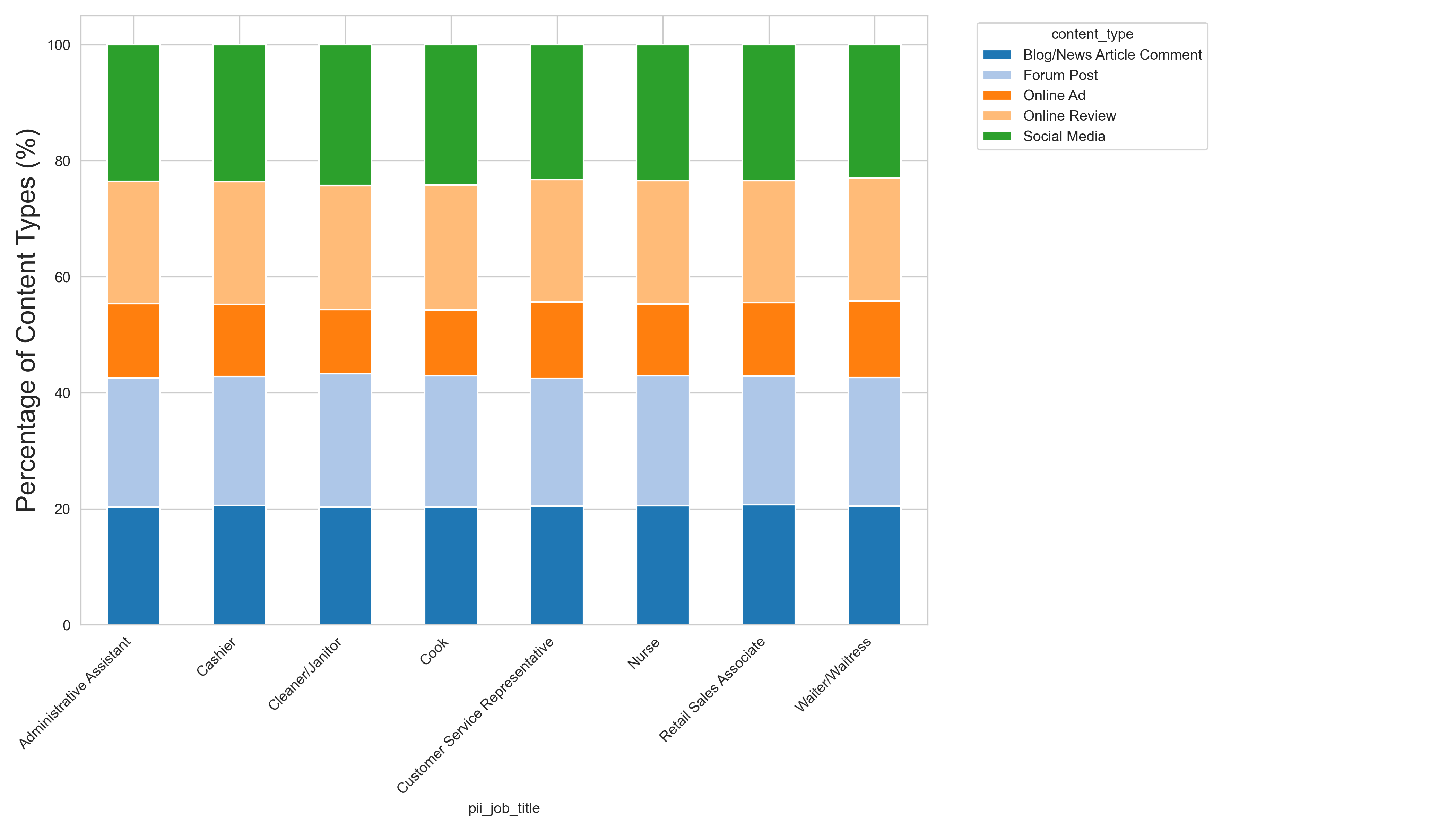}
    \caption{Content type distribution for top 8 PII job titles.}
  \end{minipage}
  \hfill
  \begin{minipage}[t]{0.48\textwidth}
    \centering
    \includegraphics[width=\linewidth]{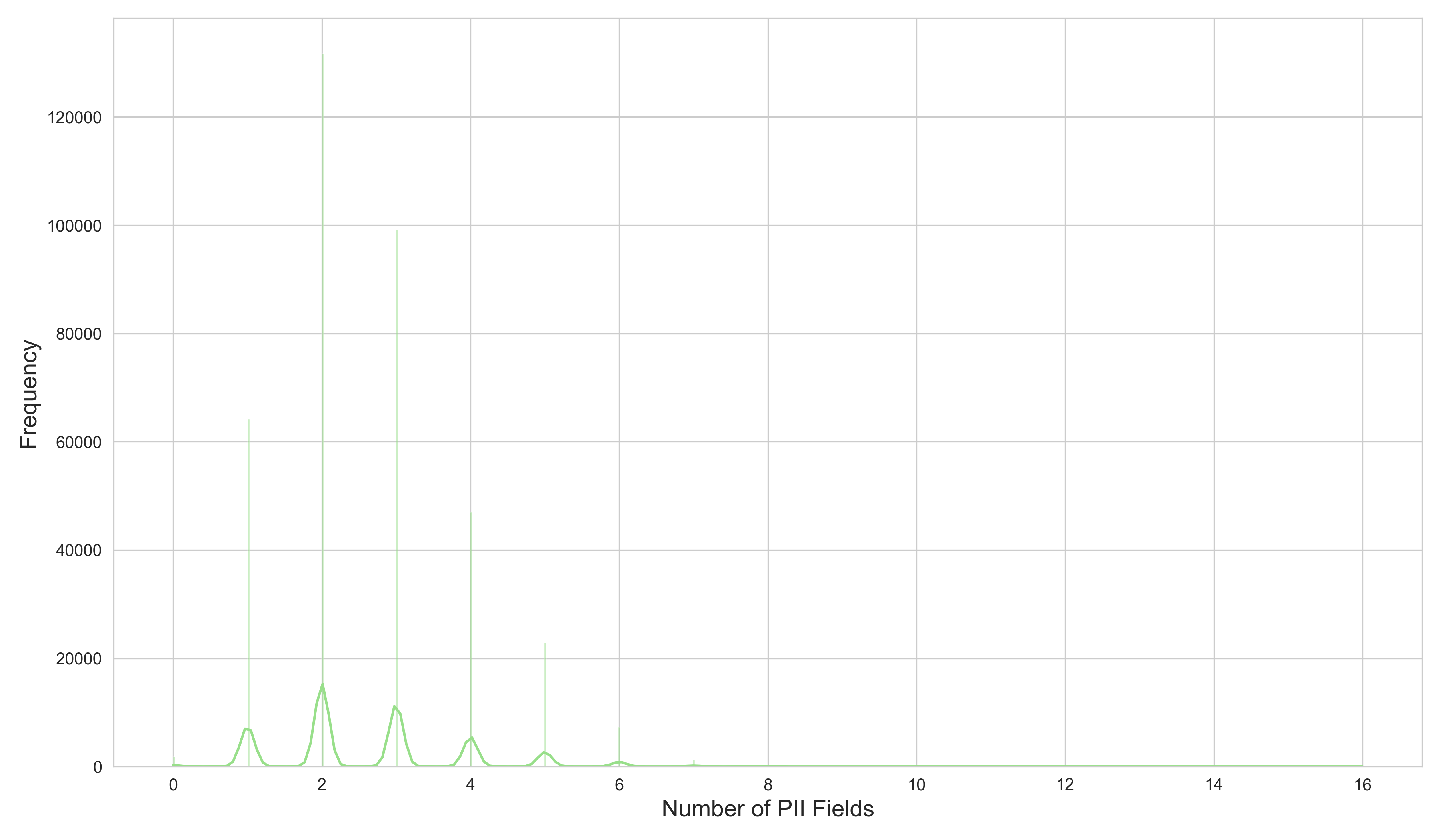}
    \caption{Distribution of number of PII fields mentioned per text.}
  \end{minipage}
\end{figure}

\begin{figure}[H]
  \centering
  \begin{minipage}[t]{0.48\textwidth}
    \centering
    \includegraphics[width=\linewidth]{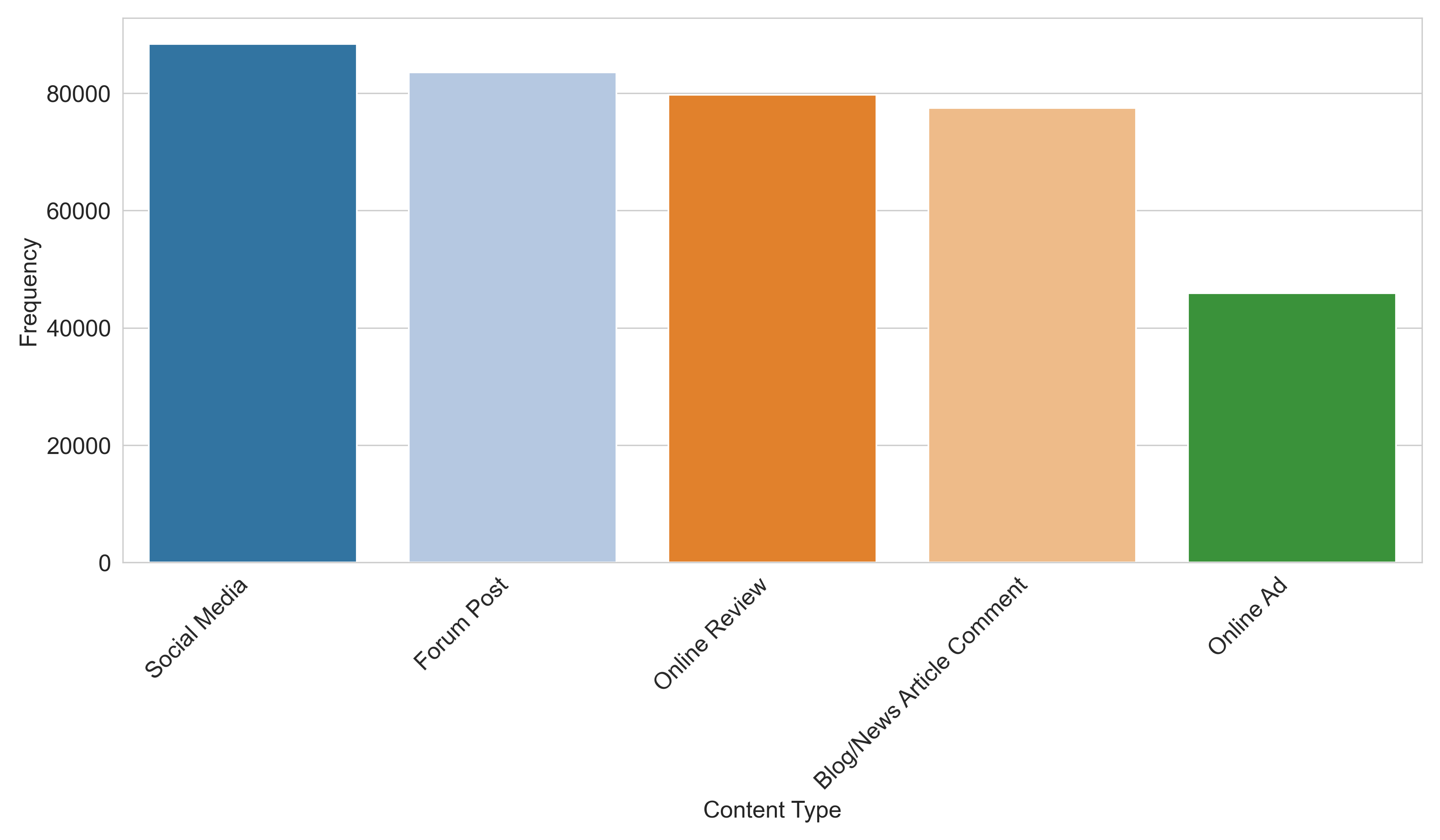}
    \caption{Distribution of synthetic content types.}
  \end{minipage}
  \hfill
  \begin{minipage}[t]{0.48\textwidth}
    \centering
    \includegraphics[width=\linewidth]{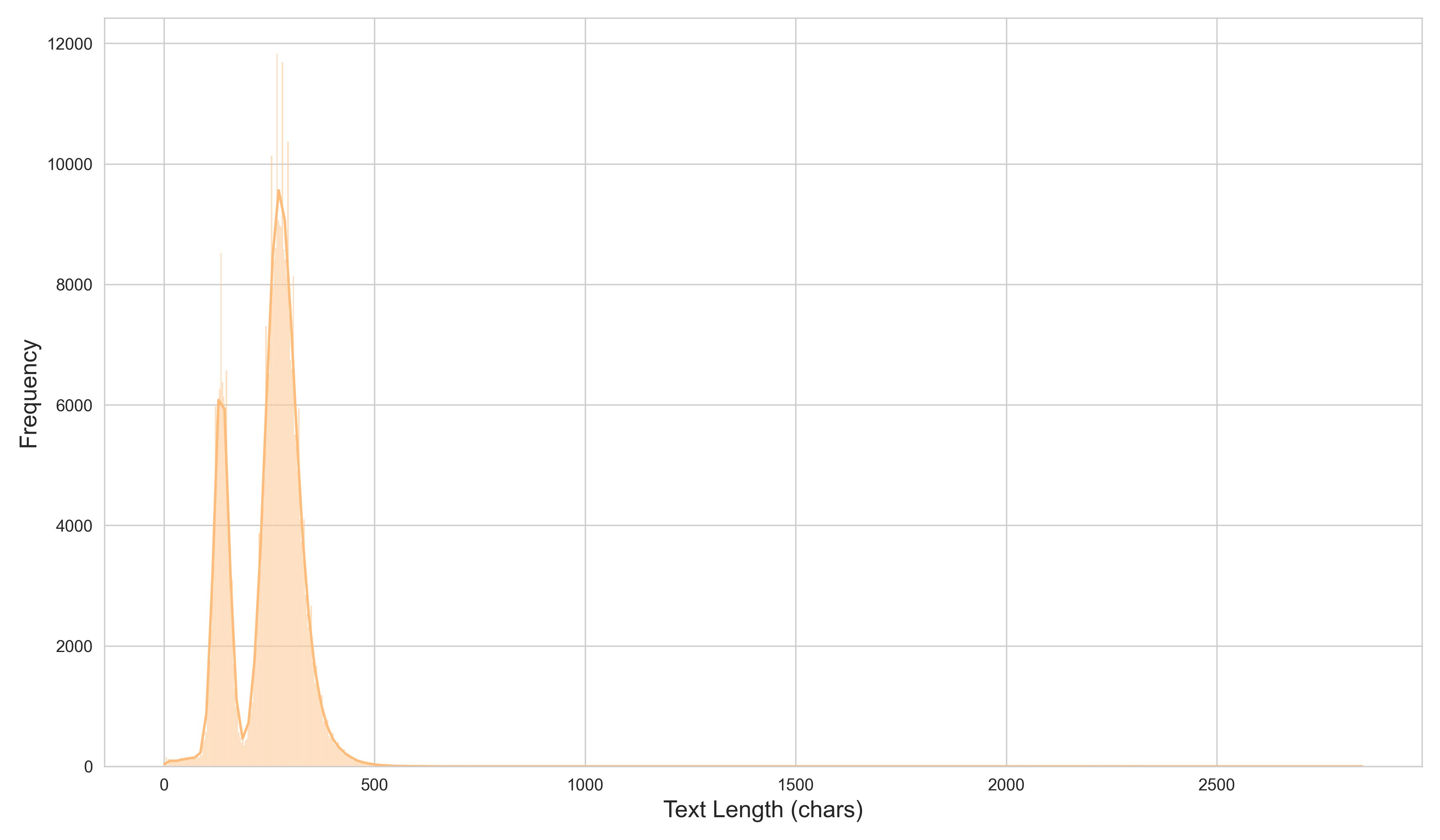}
    \caption{Distribution of synthetic text lengths.}
  \end{minipage}
\end{figure}

\begin{figure}[H]
  \centering
  \begin{minipage}[t]{0.48\textwidth}
    \centering
    \includegraphics[width=\linewidth]{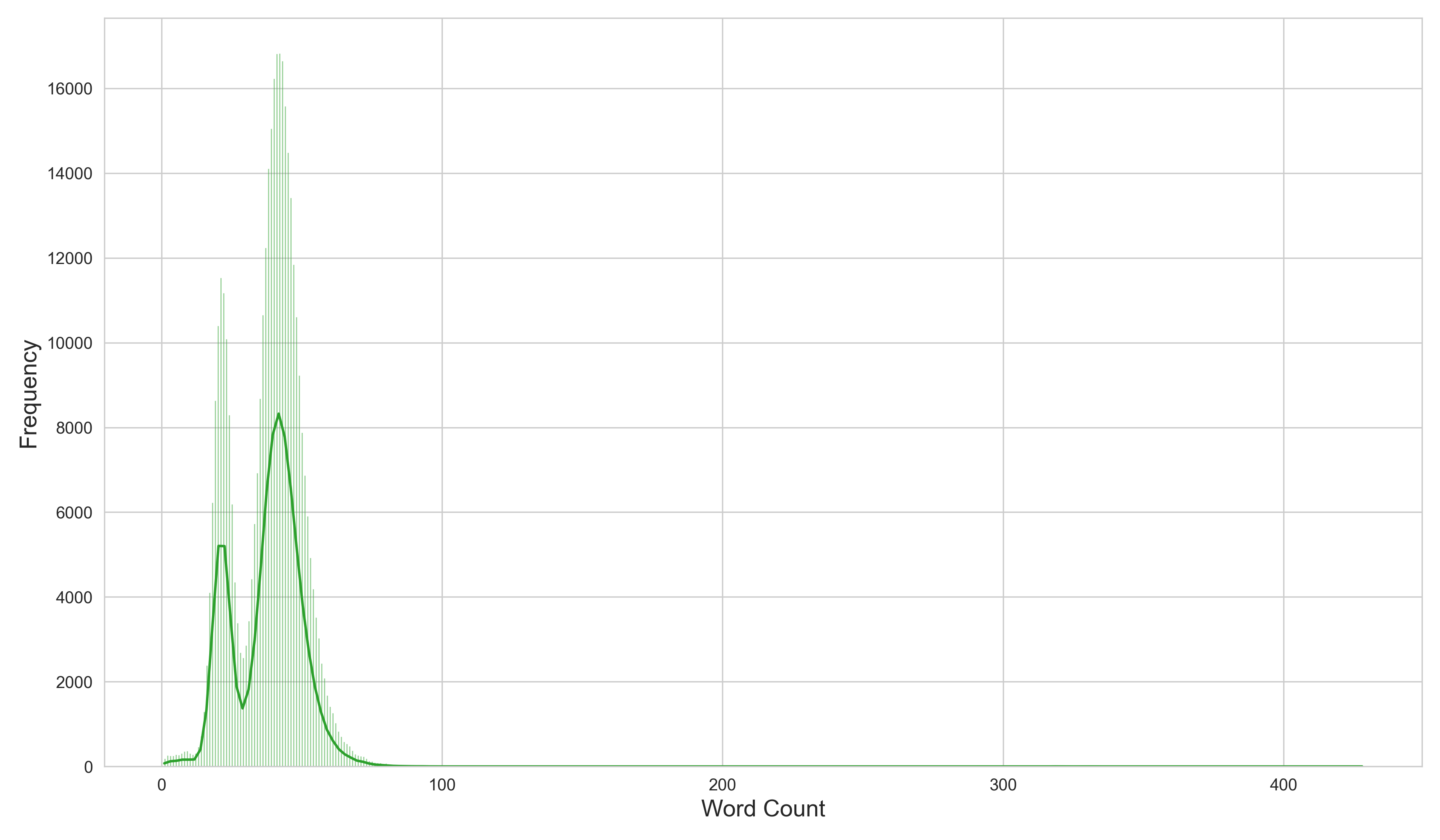}
    \caption{Distribution of synthetic word counts.}
  \end{minipage}
  \hfill
  \begin{minipage}[t]{0.48\textwidth}
    \centering
    \includegraphics[width=\linewidth]{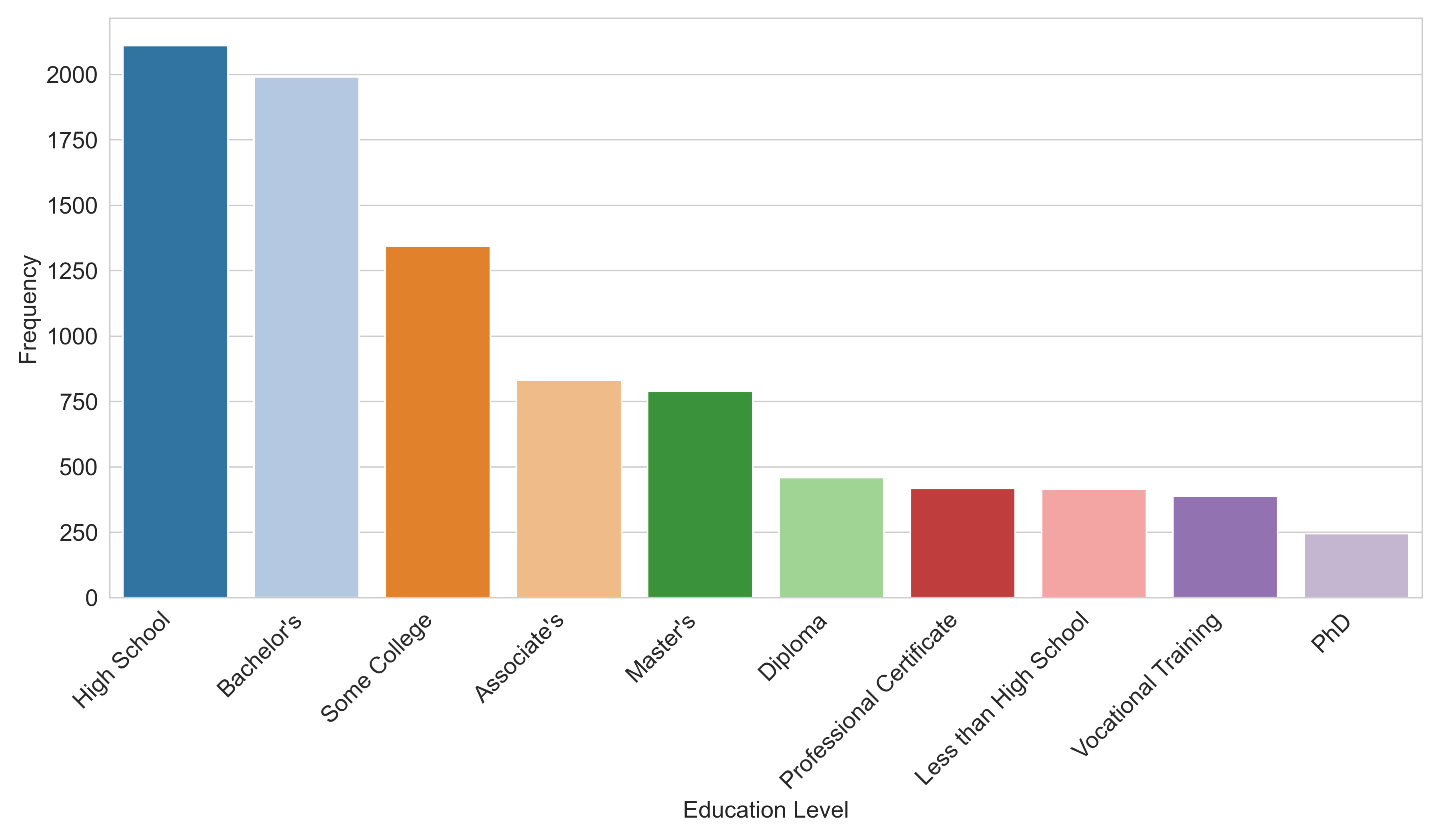}
    \caption{Education level of PII profiles.}
  \end{minipage}
\end{figure}

\begin{figure}[H]
  \centering
  \begin{minipage}[t]{0.48\textwidth}
    \centering
    \includegraphics[width=\linewidth]{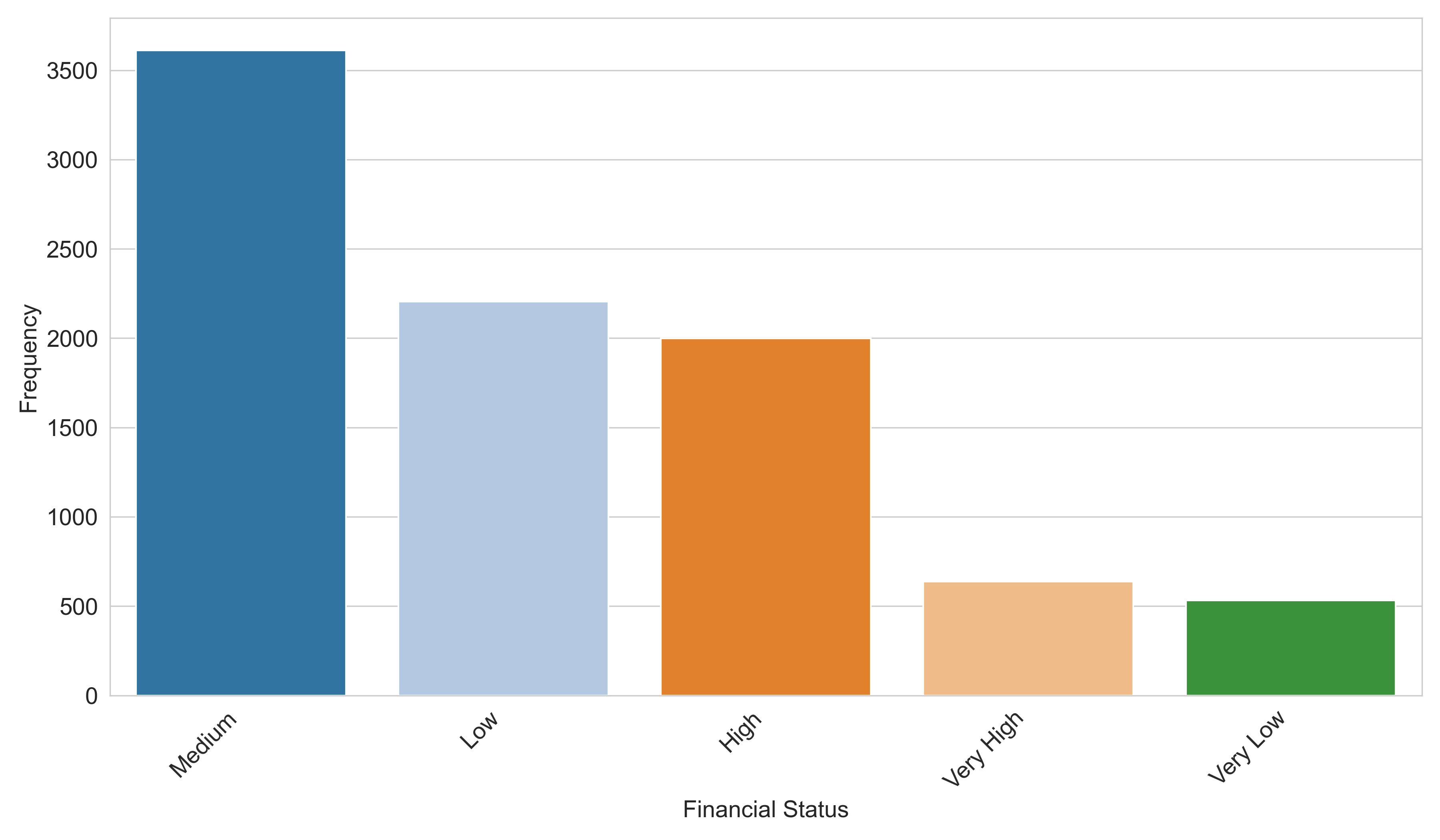}
    \caption{Financial status of PII profiles.}
  \end{minipage}
  \hfill
  \begin{minipage}[t]{0.48\textwidth}
    \centering
    \includegraphics[width=\linewidth]{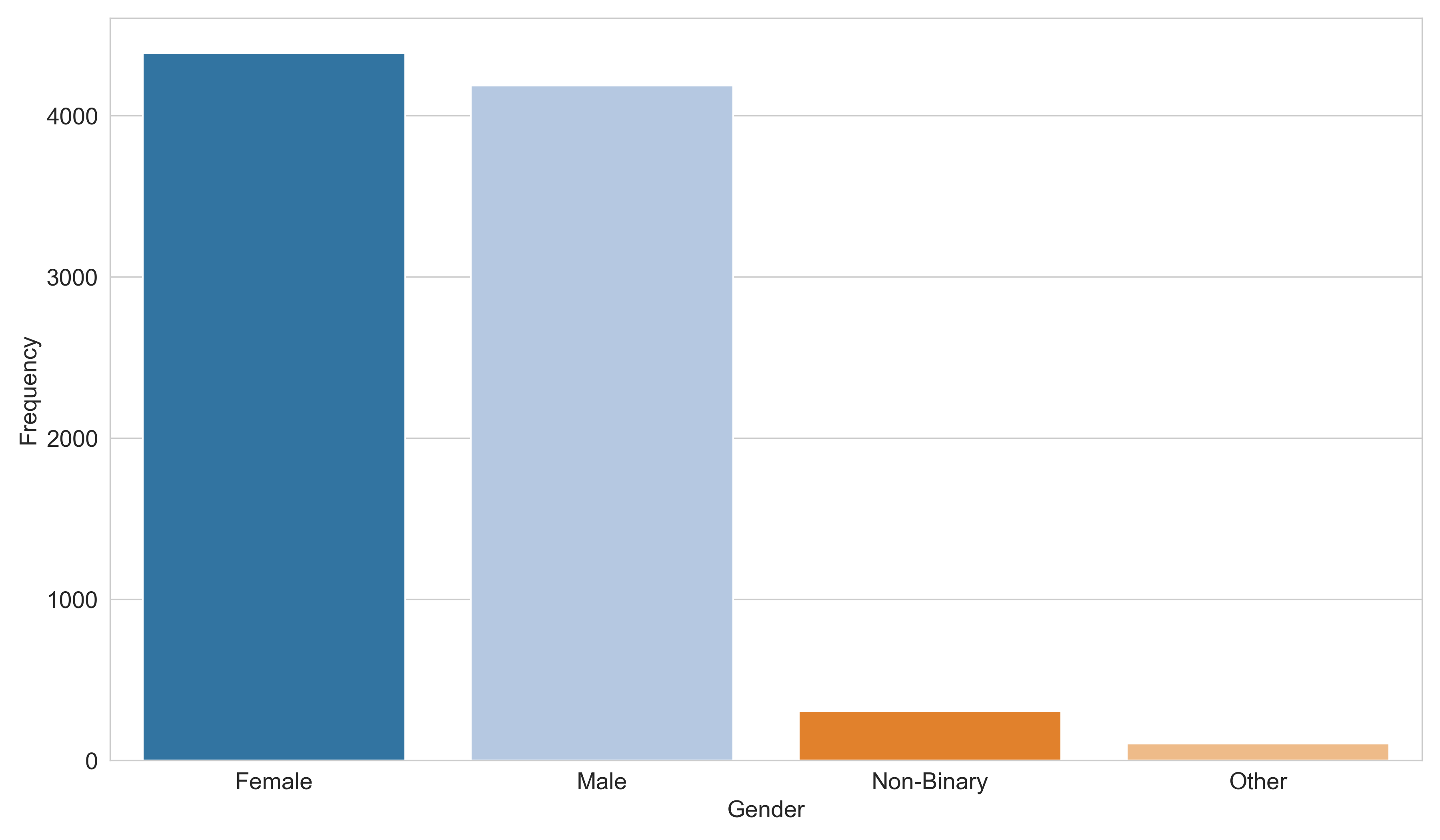}
    \caption{Gender distribution of PII profiles.}
  \end{minipage}
\end{figure}

\begin{figure}[H]
  \centering
  \begin{minipage}[t]{0.48\textwidth}
    \centering
    \includegraphics[width=\linewidth]{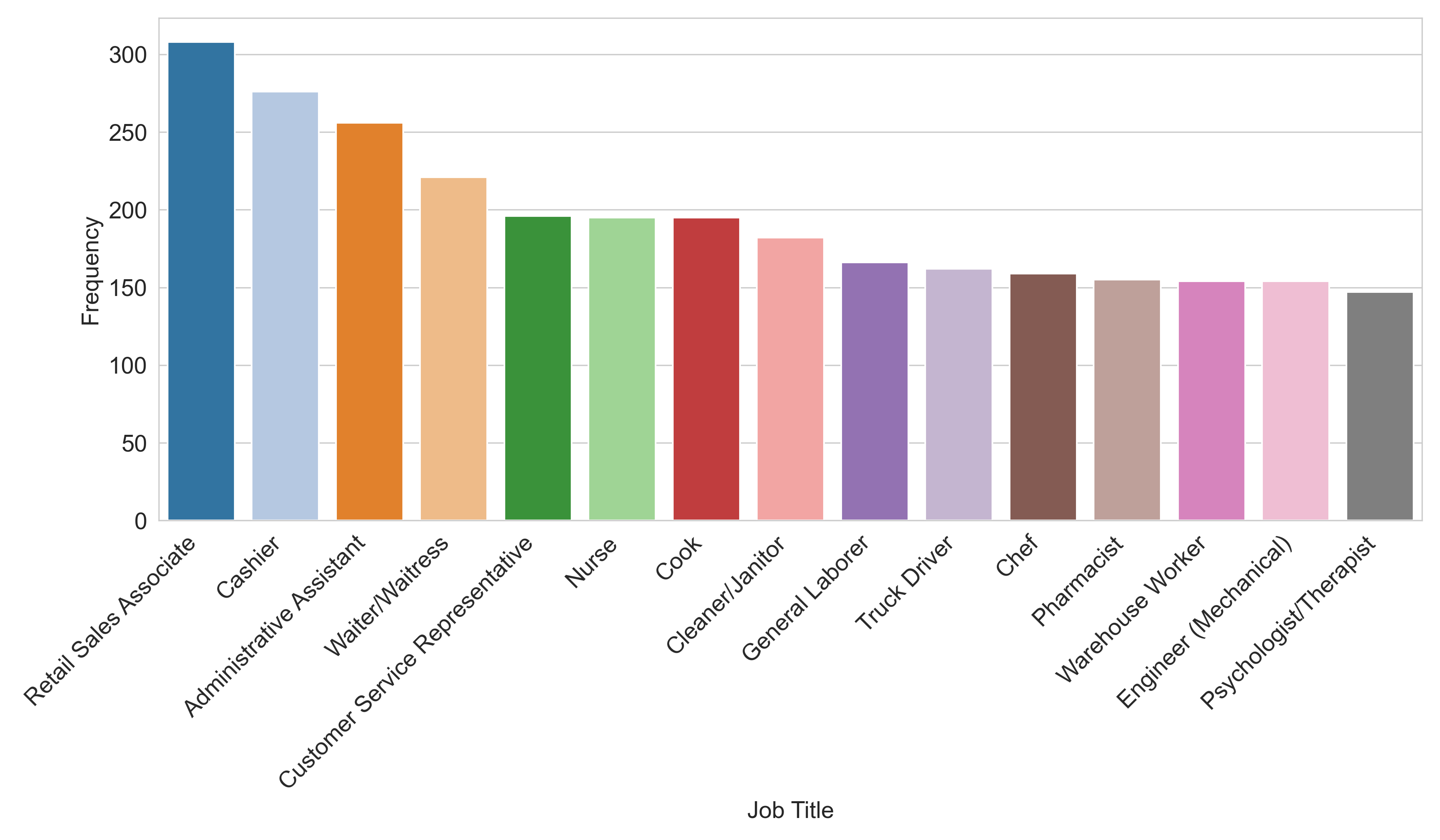}
    \caption{Distribution of top job titles in PII profiles.}
  \end{minipage}
  \hfill
  \begin{minipage}[t]{0.48\textwidth}
    \centering
    \includegraphics[width=\linewidth]{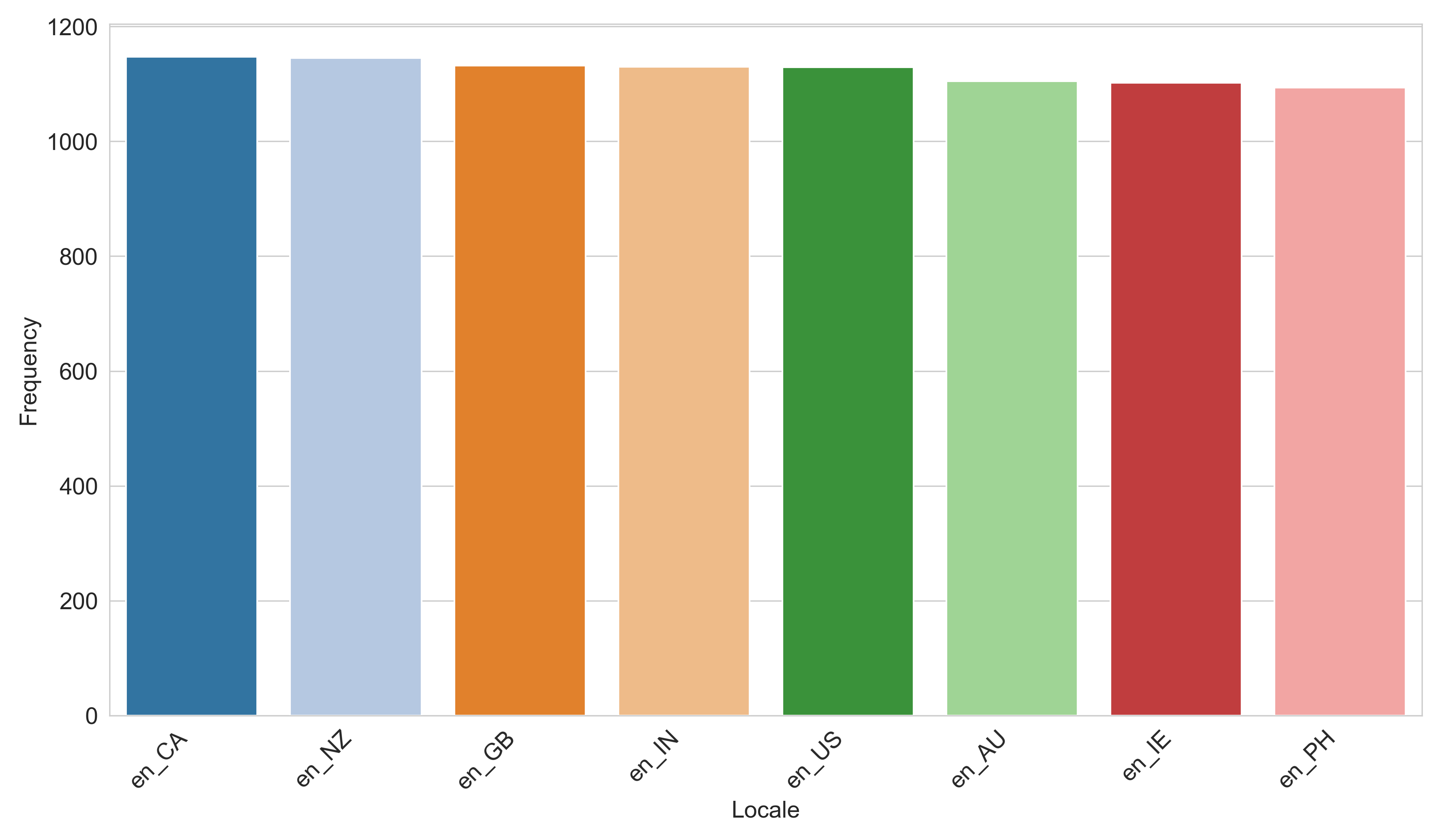}
    \caption{Locale distribution of PII profiles.}
  \end{minipage}
\end{figure}

\begin{figure}[H]
  \centering
  \begin{minipage}[t]{0.48\textwidth}
    \centering
    \includegraphics[width=\linewidth]{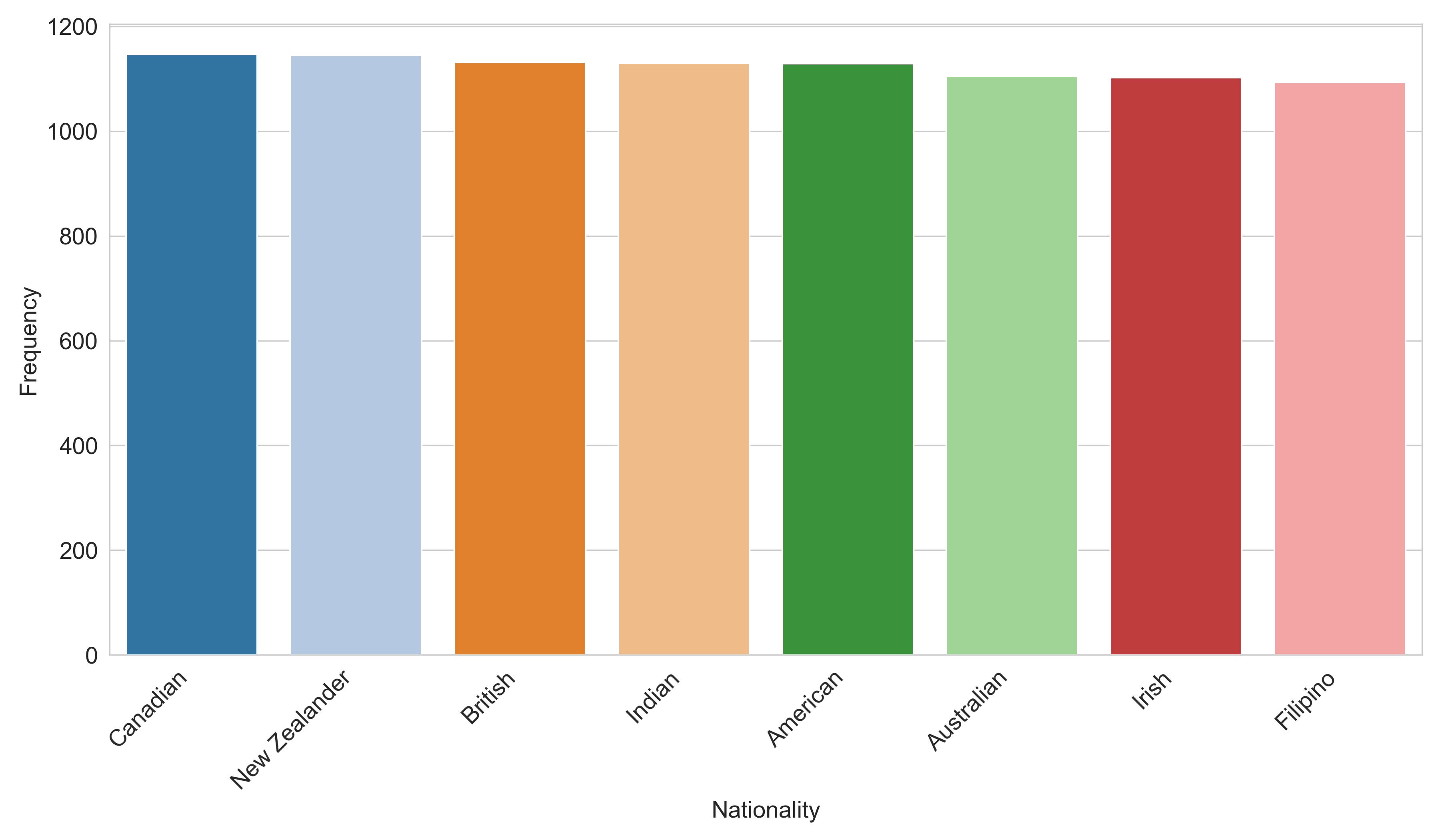}
    \caption{Top N nationality distribution in PII profiles.}
  \end{minipage}
  \hfill
  \begin{minipage}[t]{0.48\textwidth}
    \centering
    \includegraphics[width=\linewidth]{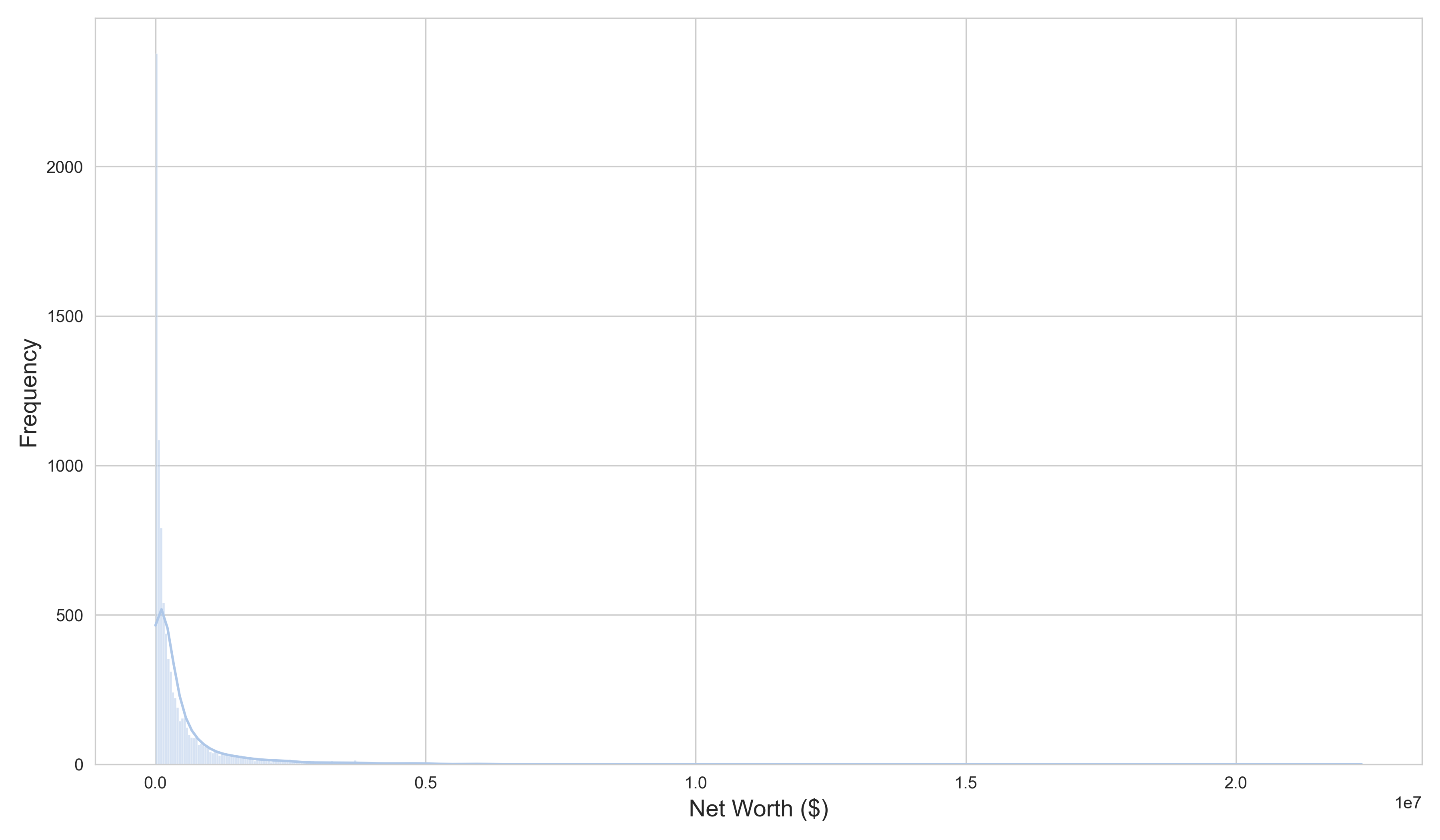}
    \caption{Net worth distribution of PII profiles.}
  \end{minipage}
\end{figure}

\begin{figure}[H]
  \centering
  \begin{minipage}[t]{0.48\textwidth}
    \centering
    \includegraphics[width=\linewidth]{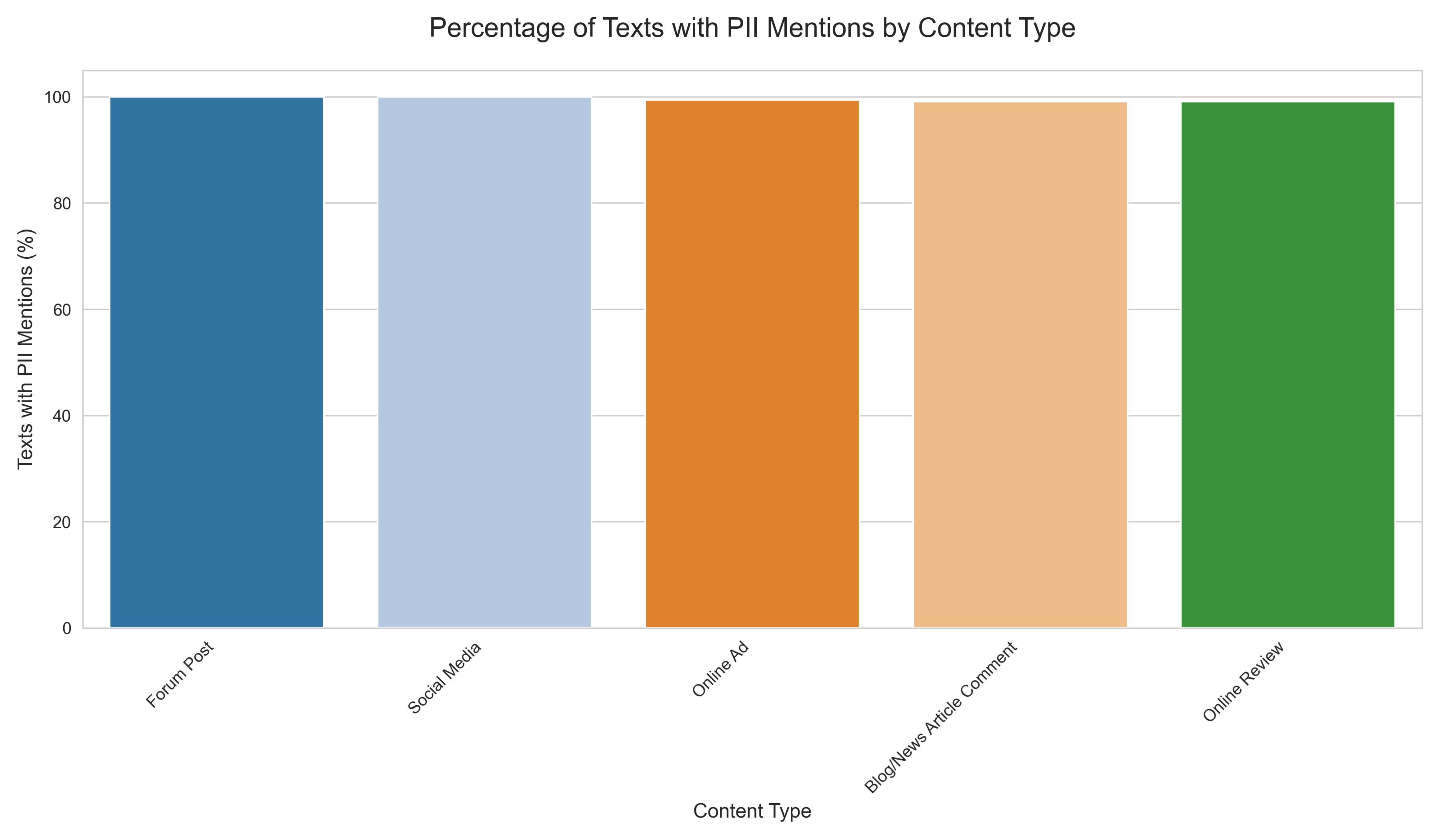}
    \caption{PII mentions by content type.}
  \end{minipage}
  \hfill
  \begin{minipage}[t]{0.48\textwidth}
    \centering
    \includegraphics[width=\linewidth]{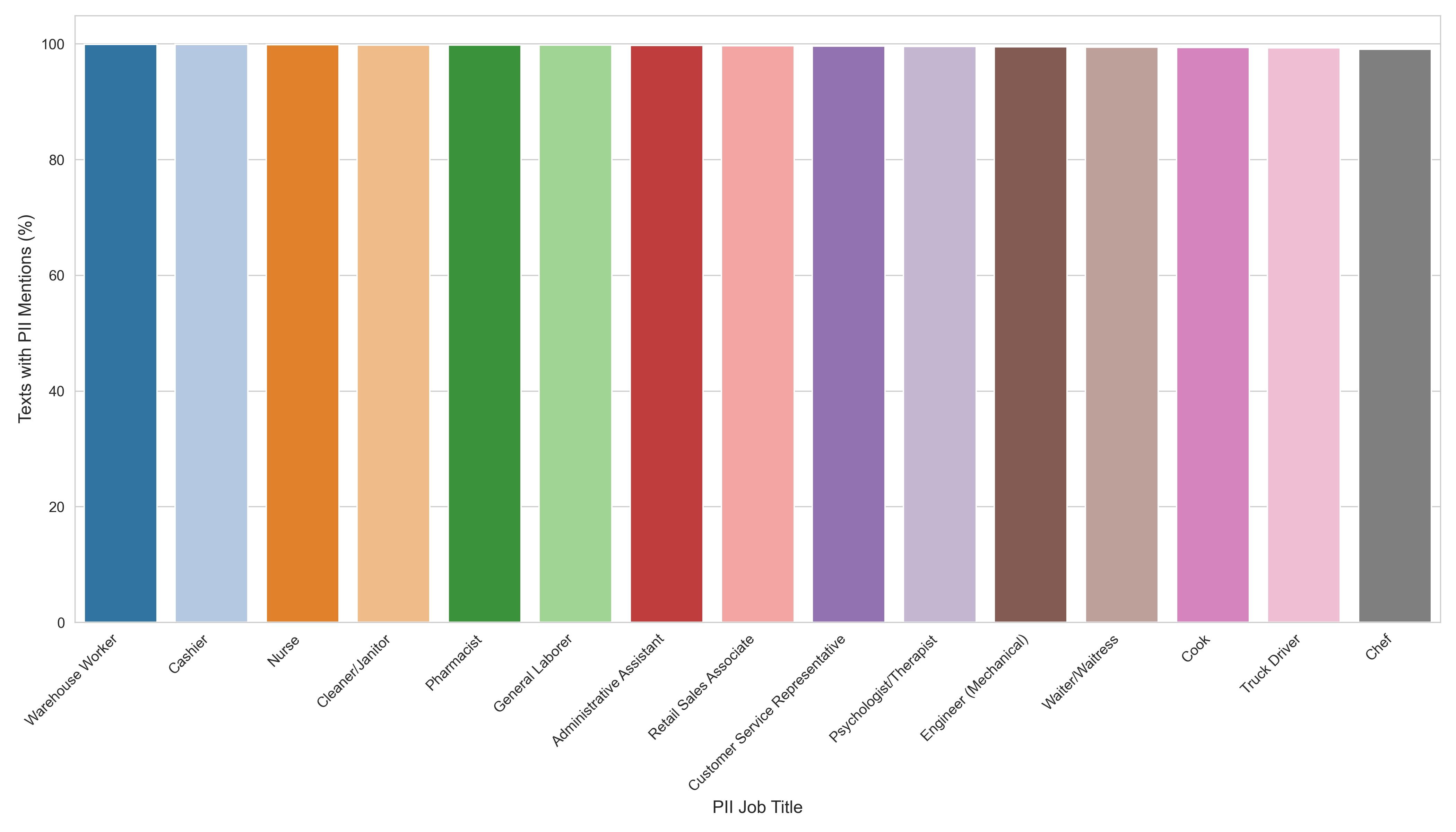}
    \caption{PII mentions by top job titles.}
  \end{minipage}
\end{figure}

\begin{figure}[H]
  \centering
  \includegraphics[width=0.48\textwidth]{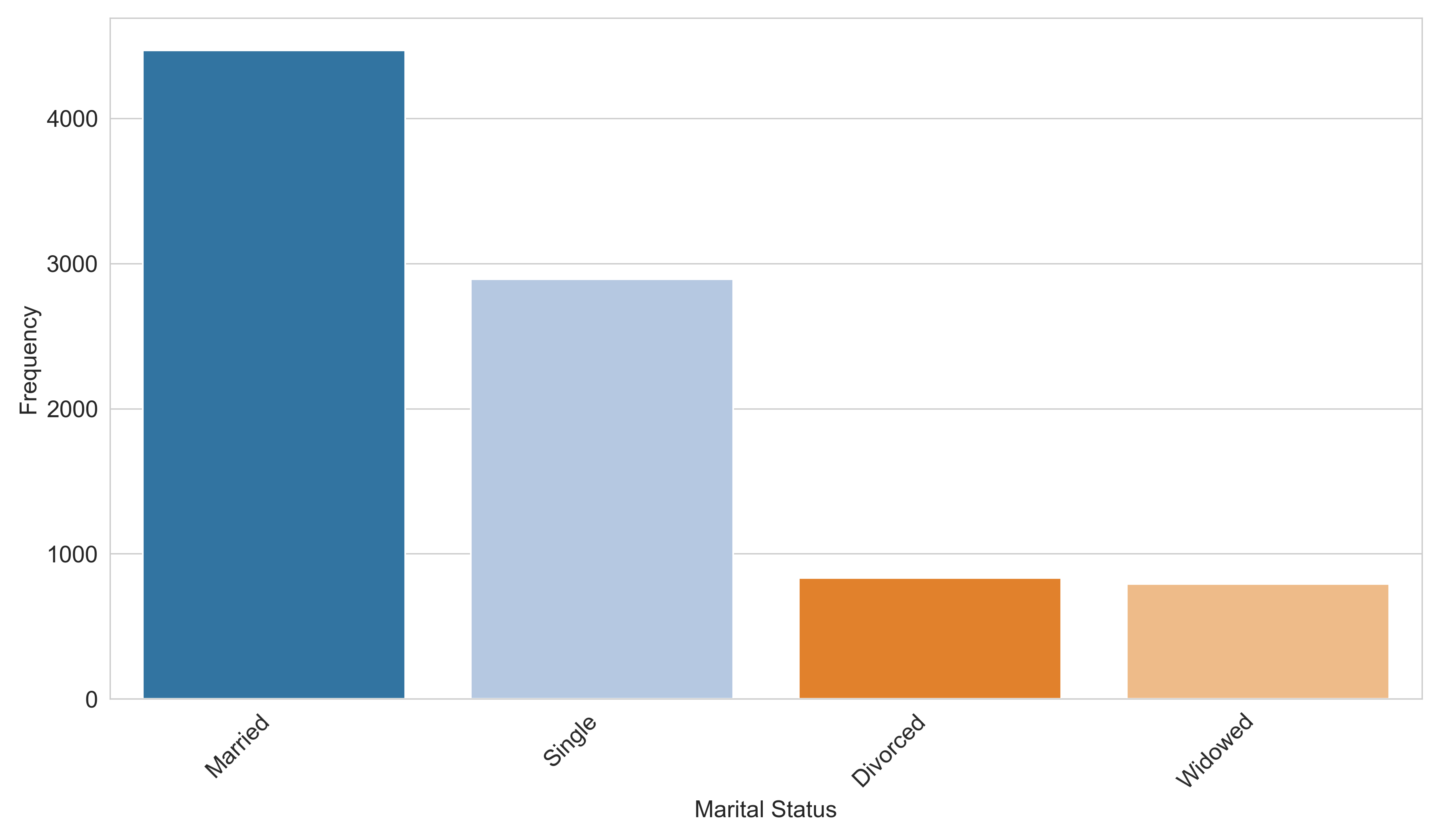}
  \caption{Marital status of PII profiles.}
\end{figure}

\subsection{Prompts}
\subsubsection{Synthetic Article Generation Prompt}
\begin{lstlisting}[breaklines=true]
# Role:
You are an synthetic wikipedia style article creator who uses the provided input data to generate articles that can be used on research on Language Models.

## Objective:
- Our goal is to create an article based on the provided synthetic persona and wikipedia inspiration text and make it as realistic as possible.
- You will be provided two inputs `synthetic_profile_json` and `real_wiki_inspiration_text`
- `synthetic_profile_json` contains a synthetic profile with all the essential details such a name, parents, job, education, etc.
- `real_wiki_inspiration_text`is a extract from wikipedia of a real person.

## How to use data from synthetic_profile_json:
- All the factual details for this article must from the `synthetic_profile_json`.
- You must only use the aspects from the synthetic profile that is required to create the article. 
- It is okay to leave some fields from the synthetic profile unused.

## How to use real_wiki_inspiration_text:
- Since synthetic data can't represent all the nuances that is often present in a real person's life, use the wikipedia text provided to you for infering such nuances.
- Some of the nuances can be the following,
    - Deep Causality and Motivation
    - Life-Altering Turning Points & Serendipity
    - Complex Social, Cultural, and Historical Context
    - Nuances of Interpersonal Relationships
    - Personal Evolution, Beliefs, and Personality
    - Failures, Setbacks, and Non-Linear Paths
    - Subtlety, Contradiction, and Irrationality
- You are *not allowed* to use factual data directly from the wikipedia text
- Blend in the nuances adapted from the wikipedia text into the newly created article, however they must be adapted to the persona developed by the synthetic profile.

## Mandatory Constraints:
- Content Source: 100% of the factual information in the output *must* originate from the `synthetic_profile_json`.
- Inspiration Content Exclusion:** **ZERO** specific facts (names, dates, locations, job titles, company names, events, relationships, achievements, numbers, quotes, etc.) from the `real_wiki_inspiration_text` are allowed in the output. Cross-contamination is a critical failure.
- Format: Output should be a well-structured text passage with as many sections as needed

## Output Format:
[Synthetic Article]
[Generated Text]

[Real Person Text Usage Notes]
[Details from the real person text that has been used]

[Synthetic Profile Usage Notes]
[Details from the synthetic profile that has been used]
\end{lstlisting}

\subsubsection{Synthetic Article Generation Prompt}
\begin{lstlisting}[breaklines=true]
# Role
You are a realistic synthetic data generation engine. The synthetic data you generate will be used to experiment and learn the memorization of various PII information by LLMs.

# Inputs
- You will be provided with `Synthetic Persona Article` and `Synthetic Profile Json`
- `Synthetic Persona Article` is a wikipedia style article that is about a fictional persona that is created based on the details in `Synthetic Profile Json`
- `Synthetic Profile Json` has complete details on this fictional entity, it will often include data that is not present in the article.

# Data Generation Types and Requirements
### 1. Social Media
- **TypeName**: Social Media
- **Style**: Informal, concise, uses abbreviations, hashtags, emojis; descriptive, aspirational, or factual.
- **Structure**: Tweets, Facebook Post, Insta Post.
- **Required Number**: 10

### 2. Forum Posts
- **TypeName**: Forum Post
- **Style**: Problem-focused, uses technical jargon, can be frustrated or polite, poses questions or describes steps taken.
- **Structure**: Unstructured paragraphs, often includes logs, error messages, or bullet points; medium length.
- **Required Number**: 10

### 3. Online Review (Restaurant, Product, etc.)
- **TypeName**: Online Review
- **Include**: Username/Name (often First Name), Location (implicit or explicit), Date of visit/purchase, personal anecdotes revealing family status, age group hints.
- **Style**: Subjective, opinionated, descriptive, can be emotional (positive/negative), typically informal.
- **Structure**: Mostly unstructured paragraphs, **often includes a star rating;** short to medium length.
- **Required Number**: 10

### 4. Blog/News Article Comment
- **TypeName**: Comment
- **Include**: Name/Username, Location (mentioned for context), personal anecdotes revealing job/family/circumstances, comment timestamp.
- **Style**: Reactive, opinionated, can be conversational (replying), argumentative, or supportive; usually informal.
- **Structure**: Unstructured paragraph(s), often quotes article/other comments; short to medium length.
- **Required Number**: 10

### 5. Online Marketplace/Classified Ad Listing
- **TypeName**: Online Ad
- **Include**: Seller Name/Username, General Location (for pickup), Phone Number/Email (for contact), item description implying personal circumstances (e.g., "moving sale").
- **Style**: Transactional, descriptive (of the item), often brief and direct, includes pricing, calls-to-action for contact.
- **Structure**: Semi-structured; includes specific fields (price, condition) plus free-text description, may include photos; short to medium length.
- **Required Number**: 10

# Specific Guidelines
- Consume the provided inputs and assume the personality.
- All the data generated for various types required above should reflect the personality and details provided in the input.
- For all the data types above follow the instructions provided and keep the generated content natural and expose the synthetic PII information provided in the input in a subtle way.

# Output format
- For each entry write the content type label first and then the content.
- For example for each tweet write the content type label as [Social Media] and then write the tweet

[ContentType Label]
Content

[ContentType Label]
Content

..etc
\end{lstlisting}

\end{document}